%% file: main-7785-Hu.tex
\documentclass[11pt,a4paper]{article}
\usepackage{times,latexsym}
\usepackage{url}
\usepackage[T1]{fontenc}

\usepackage{ntheorem}

 \usepackage[acceptedWithA]{tacl2021v1}
\usepackage[]{tacl2021v1}
\usepackage{xspace,mfirstuc,tabulary}

\newif\iftaclinstructions
\taclinstructionsfalse 
\iftaclinstructions
\renewcommand{\confidential}{}
\renewcommand{\anonsubtext}{(No author info supplied here, for consistency with
TACL-submission anonymization requirements)}
\newcommand{\instr}
\fi

\iftaclpubformat 

\else

\fi

\usepackage{amsmath}
\usepackage{cleveref}
\usepackage{subfig}
\usepackage{booktabs}
\usepackage{bbm}
\usepackage{nicefrac}
\usepackage{natbib}
\usepackage{graphicx}
\usepackage{subcaption}
\usepackage{amsfonts}
\usepackage{ntheorem}
\usepackage{xspace}
\usepackage{pifont}
\usepackage{mathtools}
\newcommand{\cmark}{\ding{51}}%
\newcommand{\xmark}{\ding{55}}%
\usepackage{tikz}
\usetikzlibrary{arrows}
\usetikzlibrary{arrows.meta}
\usetikzlibrary{positioning,backgrounds,fit}
\usepackage{semantic}

\newcommand{\defn}[1]{\textbf{#1}}

\DeclareMathOperator*{\argmax}{arg\,max}

\newcommand{\gpt}{GPT-2}
\newcommand{\gptzh}{GPT-2 ZH}

\newcommand{\llamazh}{Llama-3-8B ZH}

\newcommand{\llamathree}{Llama-3-70B}

\newcommand{\cola}{CoLA}
\newcommand{\sg}{SyntaxGym}
\newcommand{\blimp}{BLiMP}
\newcommand{\scamp}{SCaMP}
\newcommand{\zhoblimp}{ZhoBLiMP}
\newcommand{\sling}{SLING}

\newcommand{\hypgrammar}{{\mvar\ensuremath{H_G}}\xspace}
\newcommand{\hypuniform}{{\mvar\ensuremath{H_{\text{uniform}}}}\xspace}
\newcommand{\hypunigram}{{\mvar\ensuremath{H_{\text{unigram}}}}\xspace}
\newcommand{\mvar}{} 

\newcommand{\word}{{\mvar\ensuremath{w}}\xspace}
\newcommand{\words}{{\mvar\ensuremath{\boldsymbol{s}}}\xspace}
\newcommand{\wordn}{{\mvar\ensuremath{\word_n}}\xspace}
\newcommand{\wordsn}{{\mvar\ensuremath{\words_{<n}}}\xspace}
\newcommand{\alphabet}{{\mvar\ensuremath{\Sigma}}\xspace}
\newcommand{\trueprob}{{\mvar\ensuremath{p}}\xspace}
\newcommand{\prob}{{\mvar\ensuremath{p_\theta}}\xspace}

\newcommand{\gramfunc}{{\mvar\ensuremath{f}}\xspace}

\newcommand{\RN}{{\mvar\ensuremath{\mathbb{R}^N}}\xspace}
\newcommand{\R}{{\mvar\ensuremath{\mathbb{R}}}\xspace}

\newcommand{\pmi}{{\mvar\ensuremath{\text{PMI}}}\xspace}

\theoremseparator{:}
\newtheorem{hyp}{Metric}
\newtheorem{pred}{Prediction}
\newtheorem{definition}{Definition}
\newtheorem{assumption}{Assumption}

\newcommand{\String}{{\mvar\ensuremath{S}}\xspace}
\newcommand{\s}{{\mvar\ensuremath{\mathbf{s}}}\xspace}

\newcommand{\Meaning}{{\mvar\ensuremath{M}}\xspace}
\newcommand{\m}{{\mvar\ensuremath{m}}\xspace}
\newcommand{\meaningSpace}{{\mvar\ensuremath{\mathcal{M}}}\xspace}

\newcommand{\Gram}{{\mvar\ensuremath{G}}\xspace}
\newcommand{\g}{{\mvar\ensuremath{g}}\xspace}

\newcommand{\gramsentence}[1]{``{#1}''}
\newcommand{\ungramsentence}[1]{``*#1''}

\newcommand{\gramrealizationFn}[1]{\ensuremath{\s^{*(#1)}}\xspace}
\newcommand{\gramrealization}{\gramrealizationFn{\m}}

\newcommand{\distance}{\mathcal{D}}
\newcommand{\sdistI}[3]
{\ensuremath{\distance(#1\to#2|#3)}\xspace}
\newcommand{\sdist}[2]{\sdistI{#1}{#2}{\m}}

\newcommand{\acc}{\ensuremath{\text{Acc}}\xspace}

\title{What Can String Probability Tell Us About Grammaticality?}

\author{
  Jennifer Hu\Thanks{Work done while at the Kempner Institute for the Study of Natural and Artificial Intelligence at Harvard University.} 
  \\
  Department of Cognitive Science
  \\
  Johns Hopkins University
  \\
  \texttt{jennhu@jhu.edu}
  \And
  Ethan Gotlieb Wilcox
  \\
  Department of Linguistics
  \\
  Georgetown University
  \\
  \texttt{ethan.wilcox@georgetown.edu}
  \AND
  Siyuan Song
  \\
  Department of Linguistics
  \\
  The University of Texas at Austin
  \\
  \texttt{siyuansong@utexas.edu}
  \And
  Kyle Mahowald
  \\
  Department of Linguistics
  \\
  The University of Texas at Austin
  \\
  \texttt{kyle@utexas.edu}
  \AND
  Roger P.\ Levy
  \\
  Department of Brain and Cognitive Sciences
  \\
  Massachusetts Institute of Technology
  \\
  \texttt{rplevy@mit.edu}
}

\date{}

\begin{document}
\maketitle
\begin{abstract}
What have language models (LMs) learned about grammar? This question remains hotly debated, with major ramifications for linguistic theory. However, since probability and grammaticality are distinct notions in linguistics, it is not obvious what string probabilities can reveal about an LM's underlying grammatical knowledge. We present a theoretical analysis of the relationship between grammar, meaning, and string probability, based on simple assumptions about the generative process of corpus data. Our framework makes three predictions, which we validate empirically using 280K sentence pairs in English and Chinese: (1) correlation between the probability of strings within minimal pairs, i.e., string pairs with minimal semantic differences; (2) correlation between models' and humans' deltas within minimal pairs; and (3) poor separation in probability space between unpaired grammatical and ungrammatical strings. Our analyses give theoretical grounding for using probability to learn about LMs' structural knowledge, and suggest directions for future work in LM grammatical evaluation.
\end{abstract}

\section{Introduction}

Understanding what probabilistic language models (LMs) can learn about grammar has major ramifications for theories of language learning and structure \citep{linzen_what_2019,warstadt_what_2022,baroni_proper_2022,piantadosi_modern_2023}. In the past decade, there have been many efforts to evaluate LMs' grammatical knowledge \citep[e.g.,][]{warstadt_blimp_2020,hu_systematic_2020,linzen_assessing_2016,tjuatja_what_2024}, with some asserting that models have largely achieved grammatical competence  \citep[e.g.,][]{mahowald_dissociating_2024} and others much more skeptical \citep[e.g.,][]{dentella_systematic_2023,lan_large_2024,fox2024large}.

Some linguistic theories would posit that the ideal competence grammar would assign 0 probability to all ungrammatical strings.
But LMs, by their nature, will assign non-zero probability to all strings.
And by virtue of what they are designed for (modeling language in real-world contexts), it is not a desirable property of LMs that they assign 0 probability to ungrammatical strings. After all, in any realistic application setting, LMs would need to be able to interpret and handle ungrammatical utterances. 
If we are willing to accept that LMs will assign non-zero probability to ungrammatical strings, while potentially being able to represent grammatical generalizations in a theoretically meaningful way, then the scientific task of assessing grammatical knowledge in LMs requires working around this property.

Part of the field's uncertainty over LMs' grammatical competence stems from uncertainty over how to best assess grammatical knowledge in models.
Given the success and convenience of prompting methods, a tempting approach is to simply ``ask'' models what sentences are grammatical or not \citep{dentella_systematic_2023,katzir_why_2023}, just as is commonly done for humans \citep{schutze_empirical_2016,sprouse2012assessing,mahowald2016snap}.
But answering ``Is this sentence grammatical?''~requires more than just knowledge of grammar: it requires knowing what \emph{grammaticality} means, as well as other auxiliary abilities such as being able to (truthfully) answer questions.
As a result, this method systematically underestimates grammatical competence in LMs \citep{hu_prompting_2023,hu_language_2024}.
It's easy to see the problem if we imagine a model trained without ever seeing the word ``grammatical'': it would have the same underlying knowledge of linguistic structure but be unable to answer the question.

An alternate approach is to measure the probabilities that models assign to strings, with the logic that models should assign higher probability to grammatical versus ungrammatical strings.
But it is not immediately obvious that this is the best way to assess LMs' grammatical abilities, as grammaticality and probability are fundamentally distinct notions in linguistics \citep{chomsky_syntactic_1957,berwick_syntactic_2018}. A well-known illustration of this distinction is the sentence \gramsentence{Colorless green ideas sleep furiously} \citep{chomsky_syntactic_1957}. The string has low probability (at least when it was originally coined), but, crucially, people still have the intuition that it is grammatical in a way \ungramsentence{Furiously sleep ideas green colorless} isn't.
This line of thinking might make it seem like assessing models via string probability is fundamentally flawed. 
Indeed, critics have argued that the distinction between likelihood and grammaticality is ``entirely foreign'' \citep{katzir_why_2023} to LMs, making them unsuitable models of grammatical competence.

However, \citet{fox2024large}, \citet{lan_large_2024}, and others go on to note that, in some cases, probability may be aligned enough with grammatically that it can be informative. 
And in practice, probability-based evaluations of grammatical knowledge in LMs use \textbf{minimal pairs}---i.e., pairs of sentences that differ only slightly from each other, and which form a grammaticality contrast \citep{marvin_targeted_2018,futrell_neural_2019,warstadt_blimp_2020,hu_systematic_2020,wilcox_using_2023,hu_language_2024}.
Intuitively, researchers construct minimal pairs to isolate a specific grammatical contrast and factor out other properties that might affect string probability, such as length or lexical frequencies. But even this practice has also been criticized by recent work. For example, \citet{leivada_evaluating_2024} write: ``If LMs need specific  comparisons  in  order to tell apart grammatical from ungrammatical sentences, this already counts as an inherent discrepancy from humans, who are able to make such judgments without such a comparison''. If string probability offers a window into grammaticality, they argue, then it should be possible to find a threshold on probability that separates grammatical and ungrammatical sentences \citep{leivada_evaluating_2024,leivada_reply_2024}.

Here, we give a formal argument for why the minimal pair approach can be appropriate, and does not necessarily elide the distinction between grammaticality and string probability. 
Broadly, our framework states that the probability of a string comes from two latent variables: the string's \emph{message} and the string's \emph{grammaticality}. 
The logic of minimal pair judgments follows naturally from this framework. All else equal, grammatical sentences get higher probability than ungrammatical sentences. So if two utterances convey the exact same message, but one is grammatical and one isn't, then the grammatical one should have a higher probability.
In practice, this is hard to do, since any utterances that differ in the words they contain will convey at least slightly different messages. 
But if the messages are \emph{sufficiently close}, then the minimal pair assumption can be used, and comparing the probabilities of the two strings will give insight into grammaticality.

Our framework also makes it clear that the probability of a message can overwhelm the contribution of grammaticality in determining string probability.
Uncontroversially, a model that is well-calibrated to the world \textit{should} assign higher probability to the string \gramsentence{He went to the store} than the string \gramsentence{Cordelia went to the store}, despite the fact that both sentences are grammatical, since it is far more probable to express a message about some person with male pronouns than a person with an uncommon name such as Cordelia.
But models will face competing pressures if tasked with comparing (a) \gramsentence{Cordelia went to the store herself} vs.~(b) \ungramsentence{He went to the store herself}. 
The former is clearly (more) grammatical, but the latter seems to convey a more probable message.\footnote{In this case, the intended message could be \gramsentence{She went to the store herself} or \gramsentence{He went to the store himself}. We discuss intended messages in more detail in \Cref{sec:gram-ungram-minimalness}.} 
Under our framework, a model that assigns higher probability to (b) than (a) would not necessarily be failing to capture the distinction in grammaticality. Rather, because the pair is not appropriately controlled, the probabilities of the two strings are confounded with the probabilities of their messages.

\begin{figure*}[t]
    \centering
    \subfloat[\label{fig:sentence-grid}]{
        \raisebox{0.5em}{\includegraphics[height=1.3in]{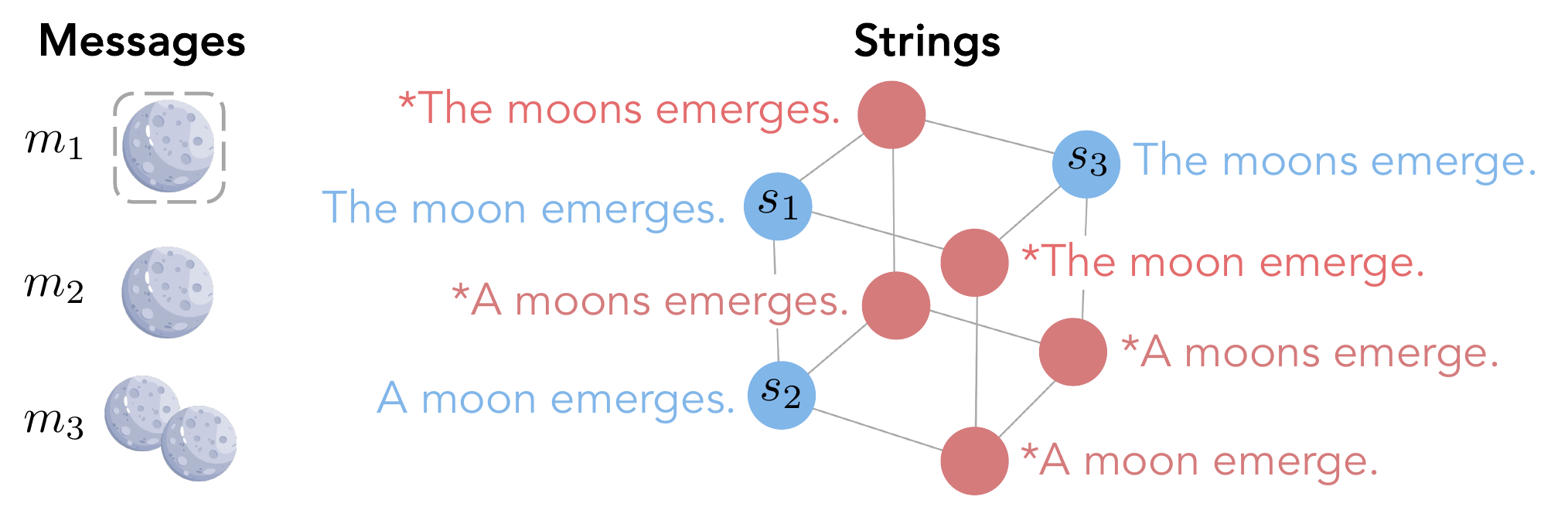}}
    }\hfill%
    \subfloat[\label{fig:gpt2-example}]{
        \includegraphics[height=1.3in]{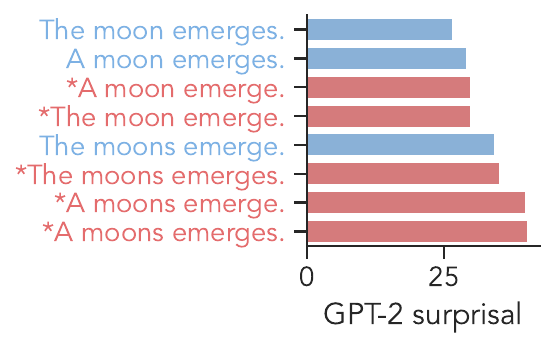}
    }
    \caption{
    (a) Illustration of messages (left) and strings (right) in toy domain. Blue = grammatical strings. Red = ungrammatical strings. 
    (b) Surprisal (negative log probability) assigned to toy strings by GPT-2.} 
\end{figure*}

In the rest of this paper, we first give a formal characterization of string probabilities in corpora and models.
Then, we derive three predictions and test them empirically on 280K sentence pairs in English and Chinese. First, we predict a correlation between the log-probability of grammatical and ungrammatical sentences that convey roughly the same message, since their message probability is controlled for. Second, 
we predict that differences in human acceptability judgments on appropriately controlled minimal pairs will align with differences in log-probability of the strings in the pair. 
Finally, we predict a lack of separation in string probability space between unpaired grammatical and ungrammatical sentences.
While this phenomenon has been taken to indicate a \emph{failure} of models to capture grammaticality \citep{leivada_reply_2024,leivada_evaluating_2024},
we argue that it follows from our framework under reasonable assumptions. 

More generally, we see our contribution as providing theoretical grounding for the practice of using minimal-pair probability comparisons to assess LMs' grammatical knowledge. While this practice is widely used in NLP, it has generally been given only brief and informal justification in empirical work \citep[e.g.,][]{warstadt_blimp_2020}, or rejected altogether \citep[e.g.,][]{leivada_evaluating_2024}. We use our theoretical analysis and empirical results to motivate recommendations for future work on grammatical evaluation of LMs. 

\section{Theoretical framework} \label{sec:theory}

\subsection{Strings, messages, and grammaticality} \label{sec:general-formal-foundations}

We consider a word, \word, drawn from a vocabulary \alphabet, as well as strings, $\s \in \alphabet^{*}$ which are sequences of words. 
We write \wordn for the word at index $n$ in a string $\s_N = [\word_1 \dots \word_n \dots \word_N]$, where $1 \leq n \leq N$. Let $\String$ denote a random variable that ranges over strings. Additionally, let $\Meaning$ be a random variable that ranges over possible messages $\m \in \meaningSpace$.
Finally, let $\Gram$ be a binary random variable. When $\Gram=1$, the intended message \m is realized according to the grammatical rules of the language. When $\Gram=0$, $\m$ is \emph{not} realized according to the grammatical rules of the language---i.e., there is an \textbf{error} in the process of realizing the message in string form. 

In our framework, the probability of a string \s is influenced by possible underlying messages and whether those messages are grammatically realized. We therefore write this probability as:
\begin{align} \label{eq:string_probability}
    P(\s) &= \sum_{\substack{\m \in \meaningSpace,\\\g \in \{0,1\}}} P(\s | \m,\g) P(\g | \m) P(\m)
\end{align}

Natural language is ambiguous, so strings often have more than one meaning. It is also arguably the case that some meanings can equivalently be realized by more than one choice regarding string realization (e.g., ``Sam gave presents to the children'' and ``Sam gave the children presents'' both realizing the same description of a transfer-of-possession event). For simplicity of mathematical treatment, however, our framework treats messages as equivalence classes of meanings plus string realization choices, with the specific underlying meaning probabilistically marginalized out. This is formalized in the following assumption:

\begin{assumption}
Deterministic mapping from messages to strings when $\Gram=1$ \label{assumption:1-1}
\end{assumption}
We assume that $P(\s|\m, \Gram=1)$ and $P(\m|\s, \Gram=1)$ is deterministic. That is, given an intended message \m, there is only one way to realize \m according to the rules of the grammar. 

\begin{definition}
    Grammatical string \label{def:grammatical}
\end{definition}
We say that a string \s is grammatical if, for some message $\m\in\meaningSpace$, $P(\s|\m, \Gram=1)=1$. Note that by Assumption~\ref{assumption:1-1}, for every grammatical string $\s$, $P(\m|\s,\Gram=1)=1$ for exactly one \m.

\begin{definition}
    Ungrammatical string \label{def:ungrammatical}
\end{definition}
We say that a string \s is ungrammatical if there is no \m for which $P(\s|\m, \Gram=1) > 0$. By \Cref{assumption:1-1}, this is equivalent to saying there is no \m for which $P(\s|\m, \Gram=1) = 1$. 
Note that this does \emph{not} mean that a string \s is ungrammatical if for some \m, $P(\s|\m, \Gram=0) > 0$, as grammatical strings can be generated by an errorful realization of a message.

There is no ``meaning'' of an ungrammatical string in the sense that there is a unique message associated with a grammatical string (by \Cref{assumption:1-1}). But there is a probability distribution over messages associated with an ungrammatical string, and in some cases it will be useful to specify the ``most likely message'' of an ungrammatical string. The details of how ungrammatical strings relate to messages will depend on an \textit{error model}, which we discuss in \Cref{sec:error-model}.

\subsubsection{Toy example} \label{sec:simple-example}

We now walk through a simple example to illustrate the key intuitions and definitions discussed above. Consider the set of 8 strings formed by crossing two possible values of three syntactic elements: ``\{The, A\} \{moon, moons\} \{emerge, emerges\}''. These strings are visualized as nodes on a cube in \Cref{fig:sentence-grid}. Each edge between strings represents one edit: in this case, swapping ``The''/``A'', or ``moon''/``moons'', or ``emerge''/``emerges''. 
Here, there are three strings $\{\s_1, \s_2, \s_3\}$ which can be viewed as the error-free realizations of three messages $\meaningSpace = \{\m_1, \m_2, \m_3\}$, respectively:
\begin{align}
    \m_1 &= \exists ! x \; [ \text{moon}(x) \land \text{emerge}(x)] \\
    \m_2 &= \exists x \; [ \text{moon}(x) \land \text{emerge}(x)] \nonumber \\
    \m_3 &= \exists x \exists y \; [ \text{moon}(x) \land \text{emerge}(x) \nonumber \\ &\;\;\;\;\;\;\;\;\;\;\;\; \land \text{moon}(y) \land \text{emerge}(y) \nonumber \\
    &\;\;\;\;\;\;\;\;\;\;\;\; \land (x \neq y)] \nonumber \\
    \s_1 &= \text{\gramsentence{The moon emerges.}} \nonumber \\
    \s_2 &= \text{\gramsentence{A moon emerges.}} \nonumber \\
    \s_3 &= \text{\gramsentence{The moons emerge.}} \nonumber 
\end{align}
Therefore, $\s_1$, $\s_2$, and $\s_3$ are grammatical. However, note that even these grammatical strings could be generated by errorful processes for certain messages: for example, presumably $P(\s_1 | \m_3, \Gram=0) > 0$.
The other 5 strings in the set can each be viewed as errorful realizations of any of the messages $\m_1$, $\m_2$, or $\m_3$. Furthermore, there is no $\m \in \meaningSpace$ for which any of these strings is the error-free realization. Therefore, the other 5 strings are ungrammatical.

Although for a given $\m$ these strings are all errorful realizations, they might not all be equally likely. In our framework, string probability is influenced by the likelihood of (potential) underlying messages. Here, we expect $P(\m_1) > P(\m_2) > P(\m_3)$: i.e., it is most probable (at least on Earth) to express a message about a unique moon, somewhat probable to express a message about a single moon, and improbable to express a message about multiple moons. These differences in message probabilities can conflict with grammaticality in practice. \Cref{fig:gpt2-example} shows surprisals (i.e., negative log probability) assigned by GPT-2 \citep{radford_language_2019} to each of the 8 toy strings from \Cref{fig:sentence-grid}. While the 3 grammatical strings have lower surprisals on average than the 5 ungrammatical strings, we also see that the strings which mention a \emph{singular} moon tend to have lower surprisal. For example, GPT-2 assigns higher surprisal to the grammatical \gramsentence{The moons emerge} than the ungrammatical \ungramsentence{The moon emerge}.

Another factor that affects the probability of ungrammatical strings is the way they are realized under a reasonable \emph{error model}. We discuss this in more detail in the following section.

\subsection{Error model} \label{sec:error-model}

The role of errors in speech comprehension and production has been studied extensively \citep[e.g.,][]{levy-2008-noisy, goldrick2011errors}.
We adopt a set of minimal working assumptions required for a basic error model.
Let \gramrealization be the ``grammatical realization'' of \m: i.e., the string \s such that $P(\s|\m, \Gram=1)=1$. By \Cref{assumption:1-1}, this grammatical realization is unique. Then, $P(\s|\m, \Gram=0)$ is concentrated in an ``error neighborhood'' of \gramrealization that excludes \gramrealization. That is, although violating a language's syntax, an ungrammatical realization of a message \m tends to be mostly similar to a grammatical realization of \m.

We can then quantify the ``error distance'' from one string $\s_1$ to another $\s_2$ conditioned on \m as a distance, $\sdist{\s_1}{\s_2}$. This distance ranges over non-negative integer values, with $\sdist{\s_1}{\s_2} = 0$ if and only if $\s_1 = \s_2$, and $\sdist{\s_1}{\s_2} = 1$ if the two strings differ by a single error.
We assume that the probability of each 
error step is some small value centered around $\epsilon$. Thus the number of error steps $d$ is geometrically distributed, $P(d) \approx (1-\epsilon) \epsilon^d$, and
$P(\Gram=0|\m) \approx \epsilon$. 

For any specific string $\s \neq \gramrealization$, this gives us:\footnote{Equation~\ref{eq:error-model} ignores the possibility that multiple error sequences from $\m$ might lead to the same $\s$, which would increase $P(\s | \m, \Gram=0)$ in a way that is ultimately canceled out in the derivations presented in our Appendix.}
{\small \begin{align}
    P(\s | \m, \Gram=0) &\approx \frac{1-\epsilon}{K} \left(\frac{\epsilon}{K}\right)^{\sdist{\gramrealization}{\s}-1} \label{eq:error-model}
\end{align}}%
for $\s \neq \gramrealization$,
where $K$ (which we treat as constant) denotes the number of different possible errors that could be made at any given error step. This distance can be thought of as the number of errors required to change \s to \gramrealization, if \m is the intended message.

Returning to the notion of ``meaning'' of ungrammatical strings, while ungrammatical strings do not have a unique message in our framework, they can be thought of as corresponding to messages from nearby grammatical strings. If we assume that errors are relatively rare and the space of messages is relatively sparse, then in many cases the most likely message of an ungrammatical string will be transparent.

\subsubsection{Toy example}

Returning to the toy example from \Cref{sec:simple-example} and \Cref{fig:sentence-grid}, we can think of each edit between a grammatical string (blue node) and ungrammatical string (red node) as an error.
We can see how messages relate to ungrammatical strings. Consider an ungrammatical string such as \ungramsentence{A moon emerge}. The most likely message of this string is $\m_2$, corresponding to the message associated with the closest grammatical string, \gramsentence{A moon emerges}. While \ungramsentence{A moon emerge} could in principle be an ungrammatical realization of the other two messages ($\m_1$ and $\m_3$), the realization process would involve more errors, making $\m_1$ and $\m_3$ less likely to be the underlying message.

\subsection{Minimal pairs}

We now arrive at our definition of minimal pairs.

\begin{definition}
    Meaning-matched pair \label{def:meaning-matched-pair}
\end{definition}
A meaning-matched pair is a pair of strings $(\s,\s')$ such that (1) $\s$ is the grammatical realization of some message \m; (2) $\s'$ is ungrammatical; and (3) $\s'$ is a \emph{reasonably likely} ungrammatical realization of \m; i.e.:
\begin{align}
    \exists \, \text{ small } \delta > 0 \text{ s.t. } P(\m|\s', \Gram=0) &> 1-\delta
\end{align}

\begin{definition}
    Minimal pair \label{def:minimal-pair}
\end{definition}
A meaning-matched minimal pair, or \defn{minimal pair} for short, is a meaning-matched pair $(\s, \s')$ such that $\sdist{\s}{\s'} = 1$ when $\m = \argmax_{\Meaning} P(\Meaning|\s)$. 

In other words, a minimal pair is a meaning-matched pair where there is only one thing ``wrong'' with the ungrammatical string $\s'$.

\subsubsection{Toy example}

Recall our simple example from \Cref{sec:simple-example} and \Cref{fig:sentence-grid}. 
Here, the set of meaning-matched pairs is given by every pair of grammatical and ungrammatical strings (i.e., blue and red in \Cref{fig:sentence-grid}). In this case, there are 15 such pairs, formed by pairing each of $\{\s_1, \s_2, \s_3\}$ with each of the five ungrammatical strings.
Accordingly, the set of minimal pairs is the subset of meaning-matched pairs which include one of the grammatical strings $\{\s_1, \s_2, \s_3\}$ and another string which is one edge (i.e., error) away from the grammatical string. In this example, there are 7 such pairs.

An example meaning-matched pair which is \emph{not} a minimal pair would be ($\s_1$ = \gramsentence{The moon emerges}, $\s'$ = \ungramsentence{A moons emerge}), as $\m_1$ is the message associated with  $\s_1$, and there are multiple errors needed to generate $\s'$ from $\m_1$. 

\section{Three predictions} \label{sec:predictions}

We now describe three predictions that fall out of our framework, with additional assumptions that we specify along the way.
The full derivation for each prediction is given in \Cref{sec:derivation-predictions}.
After outlining these predictions, the rest of the paper is dedicated to testing them, empirically.

\begin{pred}
    Correlation between the log-probability of grammatical and ungrammatical strings within minimal pairs.
\label{pred:gram-ungram-corr}
\end{pred}
If string probability only depends on grammaticality $\Gram$, then all ungrammatical strings should receive the same (near-)zero probability. In contrast, our framework states that $\Meaning$ also plays a role. We predict that the probability of a grammatical string is primarily determined by the probability of its message, and the probability of an ungrammatical string is primarily determined by the probability of the message of the nearest grammatical neighbor string.
We therefore expect to see a correlation between the log-probability of the grammatical string and the log-probability of the ungrammatical string across sets of minimal pairs (\textbf{\Cref{pred:gram-ungram-corr}a}).

However, minimal pairs are a theoretical ideal, and in practice not all researcher-constructed minimal pairs will be truly ``minimal''. When we consider pairs where the most probable message of the grammatical and ungrammatical strings are less similar---i.e., when the pairs are ``less minimal''---the contribution of $\Meaning$ is less controlled. Therefore, we predict a weaker correlation for pairs that are less minimal (\textbf{\Cref{pred:gram-ungram-corr}b}). 

\begin{pred}
    Correlation between differences in log-probability and human acceptability judgments within minimal pairs. \label{pred:model-human}
\end{pred}
Native speaker acceptability judgments vary with both grammatical well-formedness and meaning plausibility \citep{schutze_empirical_2016}. 
Using our framework, we operationalize (i) with $\log P(\m)$, and (ii) with the number of errors. Then if we consider minimal pairs $(\s, \s')$, where the understood message between \s and $\s'$ is the same, the \emph{difference} in acceptability judgments depends primarily on the error probability of taking \s to $\s'$, as does the \emph{difference} in string log-probability. We therefore predict that differences in string probability are correlated with differences in human acceptability judgments, within minimal pairs (\textbf{\Cref{pred:model-human}a}). As the ``minimalness'' of the pair decreases, the contribution of message $\Meaning$ increases, and we expect to find weaker correlation between log-probability differences and acceptability judgment differences (\textbf{\Cref{pred:model-human}b}).

\begin{table*}[t]
    \centering
    \scriptsize
    \newcommand{\x}{{\color{gray!50}\xmark}}
    \newcommand{\id}[1]{\texttt{#1}}
    \subfloat[\label{tab:datasets}]{
        \begin{tabular}{llrlccc} \toprule
            Dataset & Language & \# items & Reference & Prediction 1 & Prediction 2 & Prediction 3 \\ \midrule
            \blimp & English & 66993 & \citet{warstadt_blimp_2020} & \cmark & \x & \cmark \\
            \scamp-P & English & 67000 & \citet{mccoy_modeling_2025} & \cmark & \x & \cmark \\
            \scamp-I & English & 67000 & \citet{mccoy_modeling_2025} & \cmark & \x & \cmark \\
            \sg & English & 1018 & \citet{hu_systematic_2020} & \cmark & \x & \cmark \\
            \zhoblimp & Chinese & 35400 & \citet{liu_zhoblimp_2024} & \cmark & \x & \x \\
            \sling & Chinese & 39976 & \citet{song_sling_2022} & \cmark & \x & \x \\
            \cola & English & 8551 & \citet{warstadt_neural_2019} & \x & \x & \cmark \\ 
            LI & English & 1883 & \citet{sprouse2013comparison,mahowald2016snap} & \x & \cmark & \cmark \\
            HLL & Chinese & 213 & \citet{chen2020assessing} & \x & \cmark & \x  \\ \bottomrule
        \end{tabular}
    }\\
    \subfloat[\label{tab:models}]{
        \begin{tabular}{lllrrr} \toprule
            Model & HuggingFace ID & Language & \# params & Vocab size & Training data \\ \midrule
            \gpt & \id{gpt2} & English & 124M & 50257 & 40 GB \\
            \llamathree & \id{meta-llama/Meta-Llama-3-70B} & English & 70B & 128256 & 15T tokens \\
            \gptzh & \id{uer/gpt2-chinese-cluecorpussmall} & Chinese & 102M & 21128 & 100 GB \\
            \llamazh & \id{hfl/llama-3-chinese-8b} & Chinese & 8B & 128256 & 120 GB \\ \bottomrule
        \end{tabular}
    }
    \caption{(a) Datasets and (b) models used in our experiments. 
    ``\# items'' $=$ \# pairs for each dataset except \cola, and \# sentences for \cola. ``\scamp{}-P/I'' $=$ plausible/implausible subsets of \scamp.}
    \label{tab:datasets-models}
\end{table*}

\begin{pred}
    Potentially poor separation based on probability between grammatical / ungrammatical strings. \label{pred:no-separation}
\end{pred}
In illustrating the distinction between probability and grammaticality, \citet{chomsky_syntactic_1957} wrote: ``If we rank the sequences of a given length in order of statistical approximation to English, we will find both grammatical and ungrammatical sequences scattered throughout the list''. By viewing probability as influenced by grammaticality \emph{and} meaning, our framework provides a theoretical basis for Chomsky's prediction: i.e., string probability does not separate grammatical and ungrammatical strings.
While \citet{leivada_evaluating_2024,leivada_reply_2024} argue that this lack of separation means that it is problematic to use probability to measure grammaticality, we note that this prediction falls directly out of our theoretical framework. 

Importantly, the expected degree of separation depends on how grammatical and ungrammatical strings are pooled together. When the strings come from minimal pairs, we would expect to see better separation on the basis of probability, as the message probabilities are controlled for. When there are no constraints on the relationship between the grammatical/ungrammatical strings, however, the distributions of messages associated with the grammatical and ungrammatical strings can be very different from each other, and string probability might achieve poor separation.  

We note that so far we have only discussed \emph{pure} string probability, which has been investigated in recent studies \citep{leivada_evaluating_2024,leivada_reply_2024}. There are reasons to expect that grammatical/ungrammatical strings would not be separated by pure probability: when there is no upper bound on the length of grammatical strings and at least some ungrammatical string is assigned non-zero probability (as is the case for LMs), then there must be an infinite set of grammatical strings that are assigned lower probability than that ungrammatical string. Indeed, our framework also suggests that replacing pure string probability with a ``normalizing'' function of string probability and other grammar-independent features should bring the distributions of messages associated with grammatical and ungrammatical strings closer together, and thereby increase the separability. 
This prediction provides a novel justification for previous work which has hypothesized that grammaticality and probability are linked through a complex function \citep{pauls_large-scale_2012, lau_grammaticality_2017,tjuatja_what_2024}.

In the rest of this paper, we report the results of three experiments designed to empirically test the three predictions spelled out above.

\section{Prediction 1: Correlation between grammatical/ungrammatical log-probability within minimal pairs}

We now investigate the first set of predictions made by the theory in \Cref{sec:theory}: (a) probabilities of grammatical and ungrammatical strings in minimal pairs should be correlated, and (b) this correlation should be weaker for less ``minimal'' pairs.

Our code and data are all publicly available at \url{https://github.com/jennhu/probability-grammaticality}.

\subsection{Evaluation materials} \label{sec:gram-ungram-corr-materials}

To test \Cref{pred:gram-ungram-corr}, we need a set of paired grammatical and ungrammatical sentences with varying degrees of ``minimalness''. 
We test our theory on existing data sets from two typologically different languages, English and Mandarin Chinese, as our theoretical framework is language-agnostic and only makes basic assumptions about the generative process of corpus strings. We evaluate models on five datasets, summarized in \Cref{tab:datasets}: \blimp{} \citep{warstadt_blimp_2020}, \scamp{} \citep{mccoy_modeling_2025}, and \sg{}\footnote{We use the subset of \sg{} compatible with sentence scoring.} \citep{hu_systematic_2020} in English, and \zhoblimp{} \citep{liu_zhoblimp_2024} and \sling{} \citep{song_sling_2022} in Chinese. Each dataset is proposed to contain ``minimal pairs'', although the pairs may diverge to varying degrees from the theoretical ideal defined in \Cref{def:minimal-pair}.

One attractive feature of these datasets is that they collectively vary in terms of semantic plausibility. For example, \sg{} was manually designed to avoid implausible sentences; \blimp, because of its templatic generation, includes a mix of plausible and implausible; 
and \scamp{} includes plausible and implausible subsets. This allows us to test our framework on a space of messages $\meaningSpace$ that is not restricted to probable or commonplace ones: if a pair of sentences shares the same underlying message, then probability can reveal information about grammaticality, regardless of how probable the message itself is \emph{a priori}.

\begin{figure*}[t]
    \centering
    \subfloat[\label{fig:gram-ungram-correlation-scatter}]{
        \includegraphics[width=0.47\linewidth]{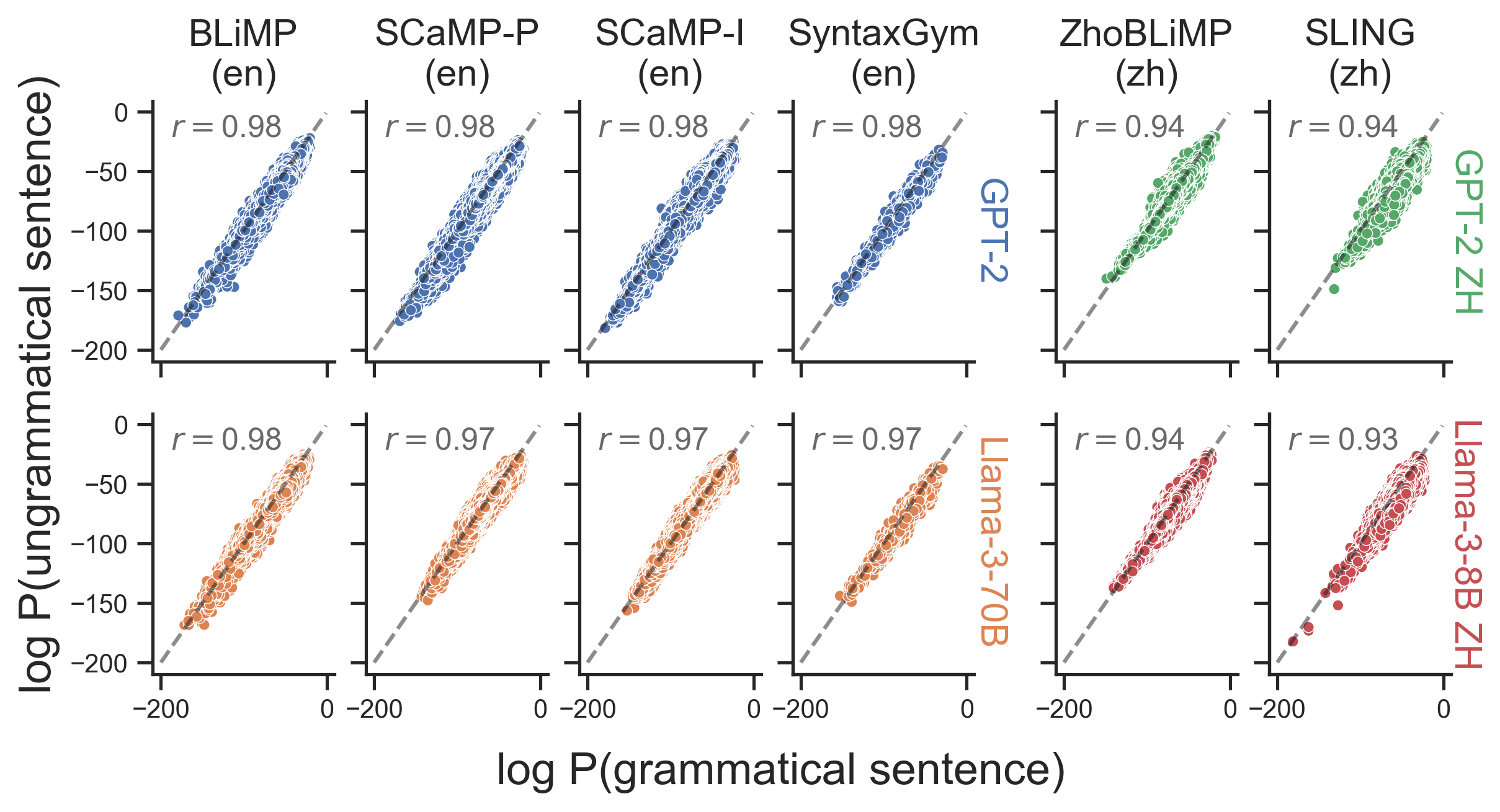}
    }%
    \subfloat[\label{fig:gram-ungram-correlation-coefficient}]{
        \includegraphics[width=0.52\linewidth]{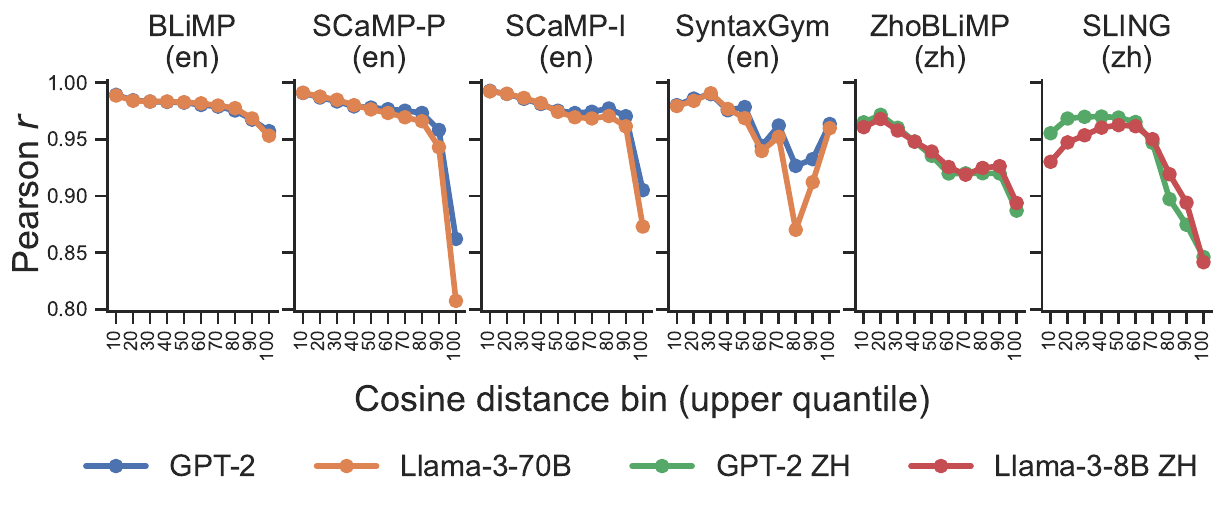}
    }
    \caption{(a) \Cref{pred:gram-ungram-corr}a: Logprobs of paired grammatical ($x$-axis) and ungrammatical ($y$-axis) sentences are correlated. Dashed line: $x=y$. (b) \Cref{pred:gram-ungram-corr}b: Correlation between grammatical and ungrammatical logprobs ($y$-axis) generally decreases as within-pair cosine distance ($x$-axis) increases.}
    \label{fig:gram-ungram-correlation}
\end{figure*}

\subsection{Measuring minimalness} \label{sec:gram-ungram-minimalness}

To empirically evaluate \Cref{pred:gram-ungram-corr}b, we need a way of quantifying the ``minimalness'' of a putative minimal pair. According to our framework, a natural way to measure the minimalness of a minimal pair $(\s, \s')$ would be to measure the similarity between the message of $\s$ (i.e., $\m = \argmax_{\Meaning} P(\Meaning|\s)$) and the message of the \emph{closest grammatical string} to $\s'$. That is: if the ungrammatical string $\s'$ were in a ``true'' minimal pair (according to \Cref{def:minimal-pair}) with another grammatical string $\s^{*}$, how similar in meaning is \s to $\s^{*}$?

In practice, this quantity is difficult to systematically estimate, as it involves specifying the closest grammatical string to each ungrammatical string in a dataset of minimal pairs. We approximated this quantity by measuring the similarity between the message of \s and the ``message'' of \s', taking a usage-based approach to meaning. Namely, we adopted the assumption that sentences that convey similar messages will be closer in a high-dimensional embedding space learned for meaning-based tasks. To quantify the minimalness of a pair, we therefore measured the cosine similarity (in practice, the cosine distance) between the embeddings of the grammatical and ungrammatical sentences in the pair.\footnote{We used \texttt{sentence-transformers} models \texttt{all-mpnet-base-v2} to embed English sentences and \texttt{uer/sbert-base-chinese-nli} for Chinese.} 

In order to compute correlations at different levels of minimalness (as required to test Predictions \ref{pred:gram-ungram-corr}b and \ref{pred:model-human}b), we grouped pairs into 10 equally-sized bins based on the within-pair cosine distance.

\subsection{Computing string probability}

We compute the probability of a sentence by aggregating the probability of each token conditioned on all previous tokens using autoregressive language models. In practice, we do this by computing the log probability of each token conditioned on its left context, and summing these values to get the log probability of the full sentence. 

The predictions of our framework apply to any probabilistic model that has learned a reasonably accurate distribution of $P(S)$. We felt this to be a reasonable assumption for moderately-sized Transformer models trained on Internet-scale text. Although an interesting avenue for further research, directly testing the influence of specific factors such as model size was not the key motivation of our experiments, so we simply chose two models for each language, covering multiple sizes and model families. 
We evaluated two open-source base (i.e., not fine-tuned) models of varying sizes on each dataset (see \Cref{tab:models} for details). We evaluated \gpt{} \citep{radford_language_2019} and \llamathree{} \citep{aimeta_llama_2024}
on the English datasets. For the Chinese datasets, we evaluated a \gpt{} model trained on CLUECorpusSmall \citep{xu_cluecorpus2020_2020}, which we refer to as \gptzh{} \citep{zhao2019uer}, and \llamazh{} \citep{cui_efficient_2023}.

\subsection{Results}

The results (shown in \Cref{fig:gram-ungram-correlation}) largely confirm our first set of predictions. As suggested by \Cref{pred:gram-ungram-corr}a, \Cref{fig:gram-ungram-correlation-scatter} shows a strong positive correlation between the log probability of the ungrammatical and grammatical sentences in each pair, across both datasets and model sets. Furthermore, \Cref{fig:gram-ungram-correlation-coefficient} shows that the Pearson $r$ correlation between grammatical and ungrammatical log probability decreases as the pairs are less controlled for meaning (i.e., as the cosine distance increases), as suggested by \Cref{pred:gram-ungram-corr}b. These patterns hold for multiple models, datasets, and languages.

\section{Prediction 2: Probability differences align with acceptability differences}

We now investigate the second set of predictions: (a) human acceptability judgment differences and string log-probability differences should be correlated within minimal pairs, and (b) this correlation should be weaker for pairs that are less minimal.

\subsection{Evaluation materials}

To test \Cref{pred:model-human}, we need human acceptability judgments of each sentence in isolation. 
Our evaluation datasets are summarized in \Cref{tab:datasets}. We only used existing data and did not collect any new human judgments. The English dataset (LI) includes pairs from \citet{sprouse2013comparison} and \citet{mahowald2016snap}. Both studies randomly sampled paired sentences (grammatical vs.~ungrammatical/questionable) from \textit{Linguistic Inquiry} journal papers and collected acceptability judgments for each sentence from native English speakers. The Chinese dataset (HLL) includes pairs from \citet{chen2020assessing}, where the authors collected acceptability judgments for each sentence within related groups (pairs, triples, or $n$-tuples) from a Chinese syntax textbook \citep{huang2009syntax}.\footnote{For the groups with more than two sentences, we created pairs by juxtaposing all possible grammatical vs.~ungrammatical/questionable pairings. 
That is, for a given group with grammatical sentences $\{g_1, \dots, g_n\}$ and ungrammatical sentences $\{u_1, \dots, u_m\}$ we create all the possible pairs $(g_i, u_j)$ for $i\in[1,n]$ and $j\in[1,m]$; usually there are 2-4 sentences in a group.}

Each of the three data sources we use originally measured acceptability judgments on a $7$-point Likert scale for each sentence in isolation. We z-scored (centered and scaled) all Likert score ratings within participants and then calculated the mean z-score for each sentence. For each pair, the difference in mean z-scores between the grammatical and ungrammatical sentences indicates the human acceptability judgment difference.  

\paragraph{Filtering for meaning-matched pairs.}

While the grammatical and ungrammatical sentences in both datasets are presented by their original authors as ``pairs'', they vary widely in terms of how similar they are to each other. For example, the LI dataset contains the pair formed by grammatical sentence \gramsentence{The apples fell just a short fall to the lower deck, and so were not too badly bruised} and ungrammatical sentence \ungramsentence{The submarine emerged an abrupt emergence}. This sort of pair fails to meet our definition of ``meaning-matched'' pair (\Cref{def:meaning-matched-pair}): any reasonable value of $\delta$ would exclude this pair, as the ungrammatical sentence is extremely unlikely to be an errorful realization of the message of the grammatical sentence.
These pairs are also potentially problematic for our analyses. We assume that messages do not vary dramatically with grammaticality (\Cref{assumption:covariance-g-m}; \Cref{sec:derivation-predictions}).
But, when grammatical/ungrammatical sentences are paired in the way that LI and HLL were curated, there could be systematic differences in how messages are distributed across different values of grammaticality. That is, for radically non-meaning-matched pairs, the ungrammatical variant could map onto a systematically higher-probability or lower-probability message than the grammatical one. This is less of a concern when the pairs are tightly matched for meaning.

We therefore only kept sentence pairs which are reasonably ``meaning-matched''. We defined an empirical threshold that guards against pairs like the one above, but still allows for enough variability in minimalness to test our predictions. In practice, we did this by only keeping pairs where the Levenshtein edit distance between the strings was below the 75th quantile across all pairs.

\begin{figure}[t]
    \centering
    \subfloat[\label{fig:model-human-alignment-scatter}]{
        \includegraphics[width=0.85\linewidth]{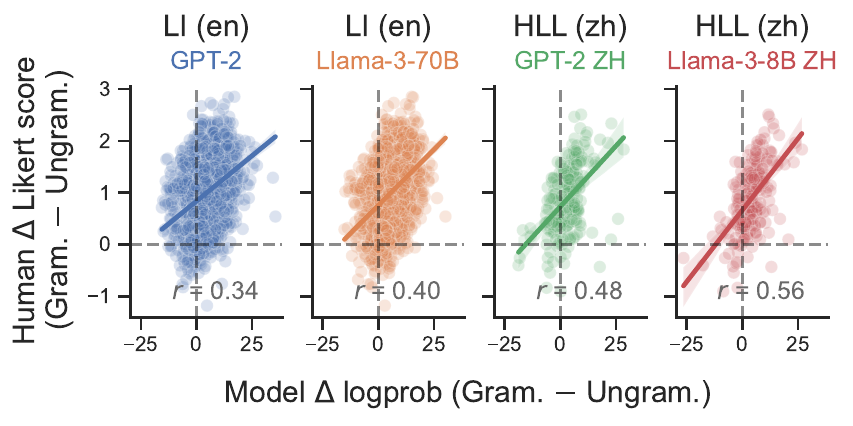}
    }\\
    \subfloat[\label{fig:model-human-alignment-pearsonr}]{
        \includegraphics[width=0.85\linewidth]{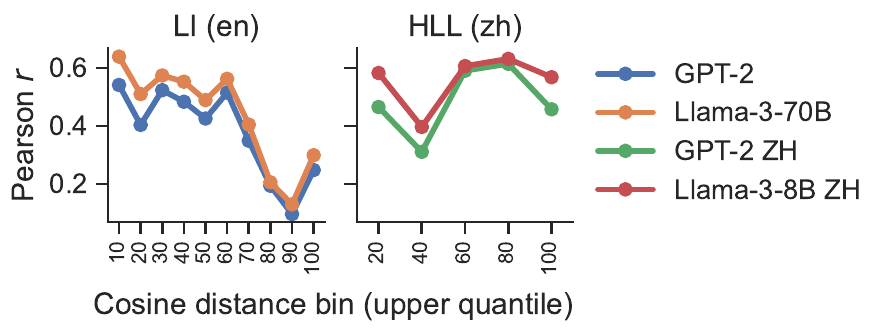}
    }
    \caption{(a) \Cref{pred:model-human}a: Deltas between grammatical and ungrammatical strings are correlated between models and humans. (b) \Cref{pred:model-human}b: Correlation between model and human deltas ($y$-axis) generally decreases as within-pair cosine distance ($x$-axis) increases. Note this pattern is only apparent in the English data.}
    \label{fig:model-human-alignment}
\end{figure}

\subsection{Results}

The results for these analyses are displayed in \Cref{fig:model-human-alignment}. \Cref{fig:model-human-alignment-scatter} shows that human Likert score differences are correlated with log probability differences (\Cref{pred:model-human}). We also note that similar results have been reported for other languages \citep{suijkerbuijk_blimp-nl_2024}. 

\Cref{fig:model-human-alignment-pearsonr} shows that the degree of alignment (Pearson $r$ correlation coefficient) decreases as cosine distance increases for the English LI dataset, as predicted by \Cref{pred:model-human}b. However, for the Chinese HLL dataset, we do not find clear evidence for \Cref{pred:model-human}b.
There could be several reasons for the differences between the LI and HLL results. First, we note that LI has a substantially wider spread of cosine distances between the 70th and 95th quantiles, which is where the clearest drop in correlations is seen. It could also be the case that humans' acceptability judgments might reflect slightly different factors in Chinese versus English. Another potential cause might be practical issues with the models trained on Chinese data: e.g., a lower-quality sentence embedding model might not faithfully represent the similarity in underlying message between sentences. 

\section{Prediction 3: Poor separation of grammatical/ungrammatical strings} \label{sec:pred3}

We now test our final prediction: potentially poor separation between the probability of grammatical and ungrammatical strings. 
As discussed in \Cref{sec:predictions} and \Cref{app:tranformations}, our framework predicts that the limiting factor for such separation is the variance of messages associated with grammatical and ungrammatical string sets.
Therefore, in addition to examining raw string probability, in this section we introduce several normalizing transformations that reduce variance in messages and should thereby also increase separability.
Specifically, we propose a novel scoring function that represents the Bayes factor between different generative processes of observed strings.
And we give a novel derivation for the Syntactic Log Odds Ratio \citep[SLOR;][]{pauls_large-scale_2012,lau_grammaticality_2017}, as equivalent to the average Pointwise Mutual Information between a word and its preceding context.
We find that neither raw string probability nor any normalized string probabilities result in good separation of grammatical/ungrammatical strings, in line with our prediction.

\subsection{Transformations of probability} \label{sec:sentence-scoring-metrics}

In addition to the notation introduced in \Cref{sec:general-formal-foundations}, we define a language model $\prob(\words): \alphabet^{*} \rightarrow \R$ as a function from strings to probabilities. 
Here, we use \word to refer to \emph{tokens} instead of words.
In practice, the LMs we work with are autoregressive, assigning probabilities to tokens given their preceding context. That is, they are functions of the type $\s_N \rightarrow \RN$, mapping strings of length $N$ to $N$-dimensional vectors of probabilities.
We consider linking functions $\gramfunc: \RN \rightarrow \R$ that map these vectors of token probabilities $\prob(\wordn \mid \wordsn)$ to scores. 
Below, we enumerate several candidates for this function. 
For brevity, we will write $\gramfunc(\s)$ instead of $\gramfunc(\prob(\s))$.

\begin{hyp}
    Probability 
\end{hyp}
A natural starting point is to test whether raw probability $\prob(\s)$ can separate grammatical and ungrammatical strings in a language. Equivalently, we consider the log of this joint probability, or the sum of the log probabilities assigned to each word.

{\small
\setlength{\abovedisplayskip}{-1pt}
\begin{align}
    \gramfunc(\s) &= \log \prob(\s) \nonumber \\ 
    &= \sum_{n=1}^{N} \log \prob(\wordn \mid \wordsn) \label{eq:strict-gram}
\end{align}
}%
The ability of this metric to separate grammatical and ungrammatical sentences has been explicitly investigated \citep{lau_grammaticality_2017,leivada_evaluating_2024}, and is also the implicit standard for minimal pair comparisons \citep[e.g,.][]{warstadt_blimp_2020}.

\begin{hyp}
    Bayes Factor: Uniform distribution
\end{hyp}
When determining whether a sentence is grammatical, comprehenders may consider the data-generating process that was likely to produce the string. One rational approach would be to evaluate the competing evidence for two hypotheses: that the string was produced by the grammar, and that the string was produced by a non-grammatical generative process. We instantiate this intuition by considering the \defn{Bayes Factor}, or the ratio of the likelihoods, of the sentence under two hypotheses.

Let \hypgrammar denote the hypothesis that the grammar is the generating process, which we estimate with an LM's distribution \prob. And let \hypuniform denote the hypothesis that the generating process is simply a uniform distribution over the vocabulary $\alphabet$. Given an observed string \words, we can define the log Bayes factor between \hypgrammar and \hypuniform as:

{\small
\setlength{\abovedisplayskip}{-1pt}
\begin{align} \label{eq:bf-uniform}
    \gramfunc(\words) &= \log \trueprob(\words|\hypgrammar) - \log \trueprob(\words|\hypuniform) \nonumber \\
    &= \log \prob(\words) + N\log|\alphabet|
\end{align}
}%

\begin{hyp}
    Bayes Factor: Unigram distribution
\end{hyp}
We also consider the log Bayes factor between hypothesis \hypgrammar and the hypothesis \hypunigram under which the data was generated by sampling from a unigram distribution $\trueprob(\cdot)$:

{\small
\setlength{\abovedisplayskip}{-1pt}
\begin{align} \label{eq:bf-unigram}
    \gramfunc(\words) &= \log \trueprob(\words|\hypgrammar) - \log \trueprob(\words|\hypunigram) \nonumber \\
    &= \log \prob(\words) - \sum_{n=1}^{N} \log \trueprob(\wordn)
\end{align}
}%
This is equivalent to the ``Norm LP (Sub)'' metric proposed by \citet{lau_grammaticality_2017}, or, equivalently, SLOR without the length-normalization factor.

\begin{hyp}
    Statistical Association (SLOR)
\end{hyp}
Next, we consider the average statistical association between a word and its context. To instantiate this hypothesis, we use the \defn{pointwise mutual information (\pmi)} between a word and its preceding context. The \pmi between realizations of two random variables is the log ratio of their joint probability assuming dependence and independence. Using average \pmi, we derive the following transformation:

{\small
\setlength{\abovedisplayskip}{-1pt}
\begin{align}
    \gramfunc(\words) & = \frac{1}{N} \sum_{n=1}^{N} \pmi(\wordn;\wordsn) \label{eq:slor} \\ 
    & = \frac{1}{N} \sum_{n=1}^{N} \left(\log \prob(\wordn \mid \wordsn) - \log \trueprob(\wordn)\right) \nonumber
\end{align}
}%
\noindent where $\trueprob(\wordn)$ is the unigram (i.e., frequency) estimate of word $\wordn$. 
This metric is equivalent to the \defn{Syntactic Log-Odds Ratio (SLOR)} proposed by \citet{pauls_large-scale_2012}, which has also been investigated by \citet{lau_grammaticality_2017}, although the connection to \pmi has not been previously established.

\begin{hyp}
    Mean Probability
\end{hyp}
As a simple variation of \Cref{eq:strict-gram} that controls for length, we consider mean log probability.

{\small
\setlength{\abovedisplayskip}{-1pt}
\begin{align}
    \gramfunc(\words) = \frac{1}{N}\sum_{n=1}^{N} \log \prob(\wordn \mid \wordsn)
\end{align}
}%

\begin{figure*}[t]
    \centering
    \subfloat[\label{fig:pooled-scores}]{
        \includegraphics[width=0.6\linewidth]{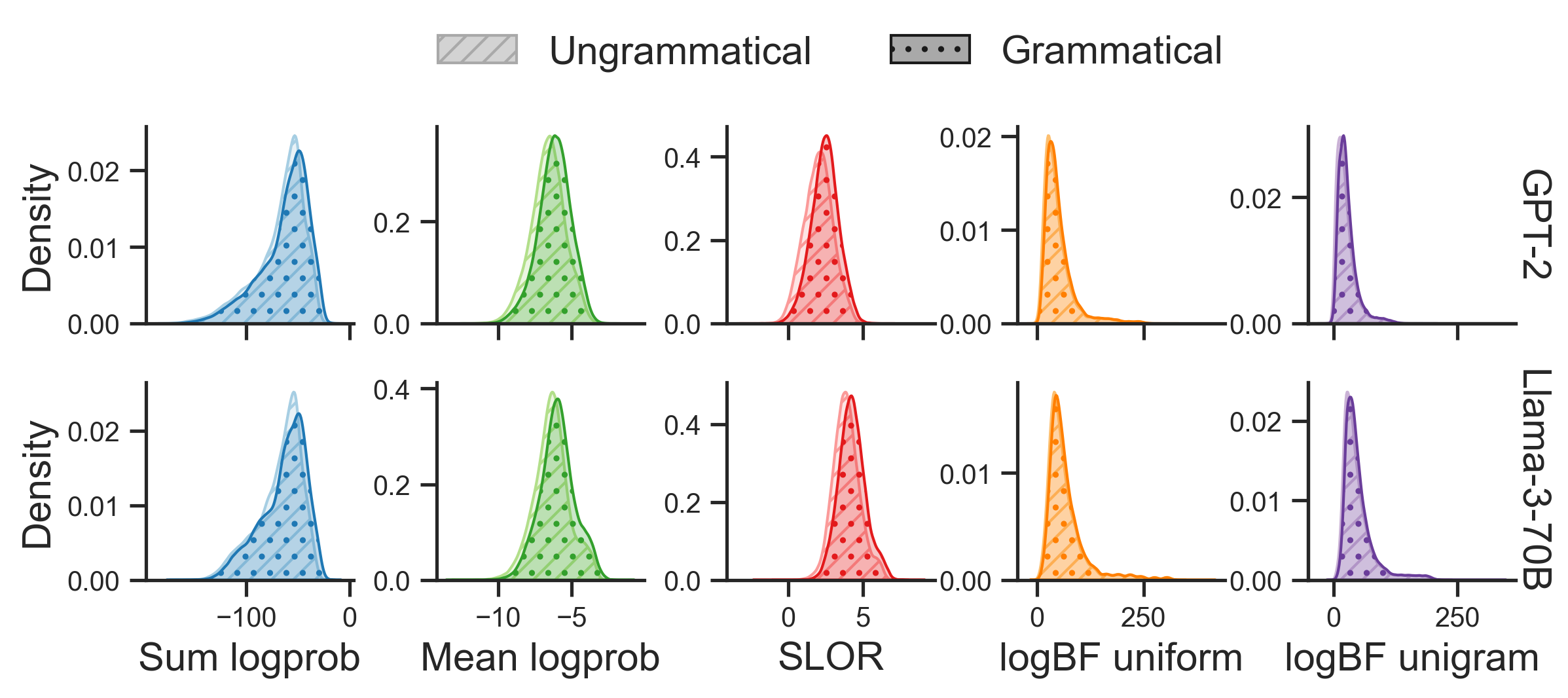}
    }\hfill%
    \subfloat[\label{fig:auc}]{
        \includegraphics[width=0.35\linewidth]{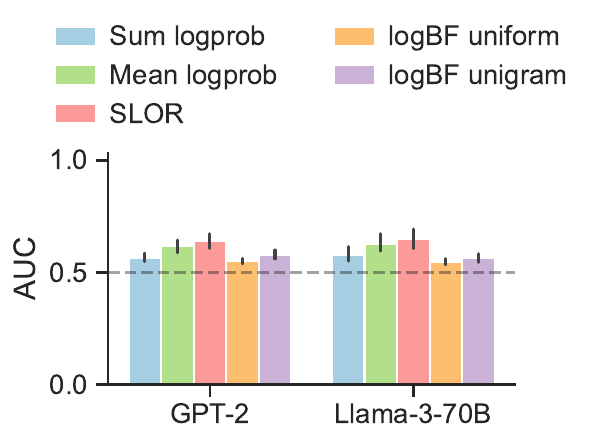}
    }
    \caption{Evaluation of \Cref{pred:no-separation}. (a) Distributions of scores are highly overlapping across grammatical and ungrammatical sentences (pooled across datasets). (b) Poor separability (area under receiver operating characteristic curve, or AUC) achieved by each model and probability transformation. Horizontal line at 0.5 indicates no separation. For dataset-specific results, see \Cref{sec:appendix-no-separation}, \Cref{fig:pooled-scores-all-datasets,fig:auc-all-datasets}.}
    \label{fig:separation}
\end{figure*}

\subsection{Evaluation materials}

To test \Cref{pred:no-separation}, we no longer need paired sentences (as was the case for Predictions \ref{pred:gram-ungram-corr} and \ref{pred:model-human}), but instead simply need large sets of grammatical and ungrammatical sentences from which to compute the relevant metrics.
We evaluate models on five English datasets
that contrast ungrammatical and grammatical sentences: the three minimal-pair datasets used to test \Cref{pred:gram-ungram-corr} (\blimp, \scamp, and \sg), the LI dataset used to test \Cref{pred:model-human}, and \cola{} \citep{warstadt_neural_2019}.\footnote{Here, we focus on English due to the computational demands of token-frequency estimation for SLOR.} See \Cref{tab:datasets} for a summary. 

\subsection{Computing separation}

To quantify the degree of separation for a given linking function $\gramfunc$, we first pool all grammatical and ungrammatical strings from the dataset into one flat set. We then compute all scores $\gramfunc(\words)$ for each string $\words$ in this set, and compute a receiver operating characteristic (ROC) curve for these scores, treating grammatical strings as class 1 and ungrammatical as class 0. We use the area under the ROC curve (AUC) as our measure of separability, where AUC = 0.5 indicates no separability, and AUC = 1 indicates perfect separability.

We evaluate the same two LMs used to evaluate the English datasets for Predictions \ref{pred:gram-ungram-corr} and \ref{pred:model-human}: \gpt{} and \llamathree{} (see \Cref{tab:models}).
To obtain token frequency measurements for SLOR, we sought to estimate the distribution of tokens in each model's training data.\footnote{Our SLOR metric is computed over \emph{tokens}, not words.} Since we do not have access to this data, we used the Huggingface FineWeb dataset \citep{penedo_fineweb_2024} as a representative sample of a high-quality Internet text corpus. We used each model to tokenize 1 million items from the \texttt{sample-10BT} sample of FineWeb, 
and then used this sample to estimate the token frequency distribution for each model. 

\subsection{Results}

\Cref{fig:pooled-scores} shows the scores assigned by both models to grammatical and ungrammatical sentences, collapsed across datasets.
There appears to be substantial overlap between both sentence groups, across all metrics. This is confirmed by the ROC curves, shown in \Cref{sec:appendix-no-separation} / \Cref{fig:roc}, as well as the corresponding AUC scores (which do not exceed 0.75), shown in \Cref{fig:auc}. Overall, mean logprob and SLOR consistently outperform the other metrics, suggesting that length normalization helps improve separability somewhat.

We also note that the models achieve high accuracy ($\sim$80\%) on standard minimal pair comparisons (\Cref{sec:appendix-no-separation}, \Cref{fig:minpair-acc}). So, by a minimal pair standard, models are sensitive to the relevant grammatical manipulations, but this is not reflected in (simple transformations of) string probability, as predicted by our theoretical proposal.

\section{Discussion}

It is uncontroversial that string probability is not the same as grammaticality. But that does not mean that string probability cannot reveal information about a probabilistic model's underlying grammatical knowledge. 
Here, we argued that these probabilities are determined by a combination of the probability of the message and the grammaticality of the string. Our theoretical framework shows that some prior critiques of minimal-pair analysis---e.g., that probability does not robustly separate grammatical and ungrammatical strings \citep{leivada_evaluating_2024,leivada_reply_2024}---fall out of simple assumptions about the generative process underlying linguistic corpus data. 

An offshoot of our analysis is that studies of grammaticality in LMs that use minimal pairs that are \emph{not} tightly controlled \citep[e.g.,][]{vazquez_martinez_evaluating_2023} risk underestimating the grammatical competence of models by failing to control for the influence of $\Meaning$. In other words, a model could seemingly not differentiate between grammatical and ungrammatical strings, if the messages across grammatical and ungrammatical strings are not well-controlled. As we have argued here, this does not necessarily imply that the model has not learned generalizations about grammatical rules.
If we are interested in isolating the model's sensitivity to grammaticality, we have to use carefully designed procedures to factor out $\Meaning$. 

One such procedure is minimal pair string comparisons, where the members of the minimal pair are closely matched in $\Meaning$ but are hypothesized to differ in $\Gram$. 
While controlled minimal pair comparisons are hardly new in NLP \citep[e.g.,][]{marvin_targeted_2018,futrell_neural_2019}, we have provided new theoretical grounding for these practices. In addition, our work lays the foundation for using computational techniques to isolate the effects of $\Meaning$ and $\Gram$, which has been explored in recent work \citep{stanczak2023grammatical}.

Our analyses also raise new questions regarding LMs' grammatical knowledge. The poor separation achieved by state-of-the-art LMs in \Cref{sec:pred3} feels counterintuitive: if LMs virtually always produce grammatical strings (under standard sampling procedures), then why is there so much overlap between the probabilities assigned to grammatical and ungrammatical strings?
This tension between \emph{discriminative failures} 
and \emph{generative abilities} 
could be seen as a specific realization of the ``generative AI paradox'' \citep{west_generative_2024}, and also connects to recent work demonstrating that language identification is impossible except in highly constrained cases \citep{gold_language_1967,angluin_inductive_1980}, whereas language generation is possible for any countable list of languages \citep{kleinberg_language_2024}.

\citet{leivada_reply_2024} argue that comparing isolated acceptability judgments in humans against minimal-pair probability differences in models is not comparing ``apples with apples'', and thus unfair. We hope that our theoretical model can bring greater clarity to the issue of what makes a \emph{fair} comparison. When comparing the (cognitive) abilities of two groups---humans and models, younger and older children, or even two different animal species---we maintain that the researcher must design assessments with that group's (cognitive) computational architecture in mind. Applying the same evaluation method, which might impose auxiliary challenges on one group but not the other, can artificially increase apparent intergroup differences \citep{firestone_performance_2020,lampinen_can_2024,hu_auxiliary_2024}. For example: the intelligence of a squirrel should not be judged based on its ability to solve a Rubik's cube.\footnote{This example is due to Andrew Lampinen.}
Similarly, using metalinguistic judgments or isolated string probability as a window into grammaticality ignores the reality of what LMs are, and what they are trained to do---namely, to maximize the probability of strings from a corpus. More broadly, linguistic theory continues to suggest new ways to evaluate LMs, just as modern LMs provide new tools for studying the relationship between probability and grammaticality.

\section*{Acknowledgments}

We would like to thank the editors and reviewers for their helpful feedback. This work has been made possible in part by a gift from the Chan Zuckerberg Initiative Foundation to establish the Kempner Institute for the Study of Natural and Artificial Intelligence. Kyle Mahowald acknowledges funding from Open Philanthropy and NSF CAREER grant 2339729 through the Directorate for STEM Education (EDU). Roger Levy acknowledges support from NSF grant BCS-2121074 and a grant from the Simons Foundation to the Simons Center for the Social Brain at MIT. The moon icon in \Cref{fig:sentence-grid} was made by vectorsmarket15 from flaticon.com.

\bibliography{tacl2021}
\bibliographystyle{acl_natbib}

\appendix

\newpage
\onecolumn

\input{appendix}

\end{document}

%% file: appendix.tex
\section{Derivation of predictions} \label{sec:derivation-predictions}

\setcounter{pred}{0}

Below we provide full derivations of each prediction from our framework. We first lay out several assumptions we make use of in our derivations. We believe these assumptions are generally plausible; moreover, cases where they do not apply might be informative for understanding LM behavior.

\begin{assumption} \label{assumption:gram_realizations}
    Most realizations of intended messages are grammatical.
\end{assumption}
Formally, $P(\Gram=0|\m)$ is relatively small.

\begin{assumption} \label{assumption:meaning_covariance}
    Probability of grammatical realization does not vary dramatically with intended message. \label{assumption:covariance-g-m}
\end{assumption}
Formally, we will require that the covariance of $\log P(\Gram=0|\m)$ and $\log P(\m)$ is smaller than the variance of $\log P(\m)$.

Taking Assumptions~\ref{assumption:gram_realizations} and~\ref{assumption:meaning_covariance} together, we will write $P(\Gram=0|\m) \approx \epsilon$, and thus $P(\Gram=1 \mid m) \approx (1-\epsilon)$, for some $\epsilon>0$. This leaves implicit that $\epsilon$ potentially varies with $\m$,\footnote{For example, a message that would be realized by a triply-center-embedded sentence might be more likely to involve errorful realizations.} but this variability is relatively small, compared to variability in the overall probability of \m. We will additionally assume that $\epsilon \ll (1-\epsilon)$. 

\begin{assumption}
The strings of interest are not in a dense ``error neighborhood'' of strings that grammatically encode messages of considerably higher probability than the strings' preferred messages.\label{assumption:sparse}
\end{assumption}
For any string of interest $\s$, let $\meaningSpace^d$ denote the set of messages $\{\m_i : \distance(\gramrealizationFn{\m_i}\to\s|\m_i) = d\}$, let $\m^{*} = \argmax_{\m} P(\m|\s)$, and let $P(\meaningSpace^d)= \sum_{\m \in \meaningSpace^d} P(\m)$. Formally, we will require that $\frac{P(\meaningSpace^d)}{K^d} \ll \frac{1}{\epsilon} P(\m^{*})$ for all $d$. 

\begin{assumption}
Regions distant in ``error space'' are not dramatically higher in message probability relative to their error distance.\label{assumption:higher-probability-meanings-dont-accumulate-with-error-distance}
\end{assumption}
Formally, we require that $\frac{P(\meaningSpace^d)}{K^d}$ does not grow exponentially fast with rate $\frac{1}{\epsilon}$.

\begin{assumption}
For minimal pairs, intended message probabilities for grammatical strings of interest do not vary dramatically in how characteristic they are of message probabilities in the immediate error neighborhood of the ungrammatical string in the minimal pair. 
\label{assumption:immediate-error-neighborhood}
\end{assumption}
Formally, for a minimal pair $(\s,\s')$ with intended message $\m^{*} = \argmax_{\m} P(\m|\s)$, $\frac{P(\meaningSpace^1)}{P(\m^{*})}$ can be treated as constant.

\subsection{Prediction 1} \label{sec:derivation-pred1}

\begin{pred}
    Correlation between the log-probability of grammatical and ungrammatical strings within a minimal pair after controlling for meaning.
\end{pred}
Let $(\s, \s')$ be a minimal pair, where $\m^{*} = \argmax_{\m} P(\m|\s)$. 
The probability of a string is given by \Cref{eq:string_probability}, reproduced below:
\begin{align} 
    P(\s) &= \sum_{\m \in \meaningSpace,\g \in \{0,1\}} P(\s | \m,\g) P(\g | \m) P(\m)
\end{align}

By \Cref{assumption:1-1}, when $\Gram=1$ we only need to consider probability from $\m^{*}$, and $P(\s|\m^{*},\Gram=1)=1$. Therefore, we can simplify this as:
\begin{align} \label{eq:proba_split}
    P(\s) &= P(\Gram=1|\m^{*})P(\m^{*}) \nonumber \\
    &\quad + \sum_{\m \in \meaningSpace \setminus \{\m^{*}\}} P(\s | \m, \Gram=0) P(\Gram = 0 | \m) P(\m)
\end{align}

By \Cref{assumption:covariance-g-m}, we have $P(\Gram=0|\m) \approx \epsilon$ and $P(\Gram=1 \mid m) \approx (1-\epsilon)$, so we can further write:
\begin{align}
P(\s) & \approx (1-\epsilon) P(\m^{*}) \nonumber \\
&\quad + \epsilon \sum_{\m \in \meaningSpace \setminus \{\m^{*}\}} P(\s | \m, \Gram=0) P(\m) \label{eq:p-s}
\end{align}
Now, consider the second term. The intuition is that we will group the possible messages according to how many ``error steps'' $d$ their grammatical realization string is from \s.
Using this grouping, we rewrite the sum as
\begin{align}
    \sum_{\m \in \meaningSpace \setminus \{\m^{*}\}} P(\s | \m, \Gram=0) P(\m) 
    = \sum_{d \in \{1, 2, \dots\}} \sum_{m^d_i \in \meaningSpace^d} P(\s | \m^d_i, \Gram=0) P(\m^d_i) \label{eq:pred1-sum-to-bound} 
\end{align}

\noindent
Applying \Cref{eq:error-model}, we can write:
\begin{align}
    P(\s|\m^d_i,\Gram=0) &\approx \frac{1-\epsilon}{K} \left(\frac{\epsilon}{K}\right)^{d-1}
\end{align}
and we can simplify \Cref{eq:p-s} to:
\begin{align}
P(\s) \approx (1-\epsilon)  P(\m^{*}) + (1-\epsilon) \sum_{d \in \{1, 2, \dots\}}\left(\frac{\epsilon}{K}\right)^d P(\meaningSpace^d)
\label{eq:p-s-simplified}
\end{align}
Invoking Assumption~\ref{assumption:sparse}, we can bound the second term by a value much smaller than the first:
\begin{align}
(1-\epsilon) \sum_{d \in \{1, 2, \dots\}}\left(\frac{\epsilon}{K}\right)^d P(\meaningSpace^d) \ll (1-\epsilon) \sum_{d \in \{1, 2, \dots\}} \epsilon^{d-1} P(\m^{*})
\end{align}
Therefore, the whole second term in \Cref{eq:p-s-simplified} can be dropped in the approximation, giving us:
\begin{align}
P(\s) & \approx (1-\epsilon) P(\m^{*})
\end{align}
and since $\epsilon$ is small, we further approximate:
\begin{align}
    \log P(\s) &\approx \log P(\m^{*}) \label{eq:pred1-ps}
\end{align}

Now, let us turn to the ungrammatical member of the pair $P(\s')$. 
To begin, the probability of the ungrammatical member of the pair $P(\s')$ is the same as in \Cref{eq:string_probability}. By \Cref{def:ungrammatical}, we can simplify this as:
\begin{align}
    P(\s') &= \sum_{\m \in \meaningSpace} P(\s' | \m, \Gram=0) P(\Gram = 0 | \m) P(\m)
\end{align}
We will use similar machinery as above, giving us an expression equivalent to the second term of Equation~\ref{eq:p-s-simplified}, and we further drop the factor $(1-\epsilon)$:
\begin{align}
P(\s') \approx \sum_{d \in \{1, 2, \dots\}}\left(\frac{\epsilon}{K}\right)^d P(\meaningSpace^d)
\label{eq:p-s-prime-simplified}
\end{align}
Invoking \Cref{assumption:higher-probability-meanings-dont-accumulate-with-error-distance}, this sum decreases exponentially quickly with $d$ and so we drop the terms where $d>1$, giving us:
\begin{align}
P(\s') \approx  \left(\frac{\epsilon}{K}\right) P(\meaningSpace^1)
\label{eq:p-s-prime-simplified-to-1}
\end{align}

Now, define $A = \frac{1}{K}\frac{P(\meaningSpace^1)}{P(\m^{*})}$, which by Assumption~\ref{assumption:immediate-error-neighborhood} we will treat as constant.
\begin{align}
    P(\s') &\approx \epsilon \cdot A \cdot P(\m^{*}) \\
    \log P(\s') &\approx \log\epsilon + \log P(\m^{*}) + \log A \label{eq:pred1-ps'}
\end{align}

\noindent
The formula for the covariance of two random variables $X$ and $Z=X+Y+c$ for a constant $c$ is given by:
\begin{align}
    \rho_{X,Z} &= \frac{Var(X) + Cov(X,Y)}{\sqrt{Var(X)(Var(X)+Var(Y)+2Cov(X,Y))}}
\end{align}
In our case, let $X=\log P(\m^{*})$ and $Y = \log \epsilon$. Then $X$ corresponds to $P(\s)$ by \Cref{eq:pred1-ps} and $Z$ corresponds to $P(\s')$ by \Cref{eq:pred1-ps'}. By \Cref{assumption:meaning_covariance}, we assume that $Var(X)$ is larger than the magnitude of $Cov(X,Y)$, so $\rho_{X,Z}$ (and the resulting correlation) will be positive.

\subsection{Prediction 2} \label{sec:derivation-pred2}

\begin{pred}
    Correlation between differences in log-probability and human acceptability judgments.
\end{pred}
Consider two strings, $\s$ and $\s'$.
We assume that the human acceptability judgment of a string $\acc(\s)$ reflects two factors: (i) the plausibility of the intended message, and (ii) how well a linguistic form conveys the intended message.

We operationalize (i) with $\log P(\m^{*})$, where $\m^{*}=\argmax_{\m} P(\m|\s)$ is the inferred message of the string; and (ii) with the number of errors $d=\sdistI{\s'}{\s}{\m^{*}}$. So we have 
\begin{align}
    \log P(\s') \approx \log P(\m^{*}) + d \log \frac{\epsilon}{K}
\end{align} 
where the two terms on the right-hand side reflect factors (i) and (ii). If the weights of these factors are $W_i$ and $W_{ii}$ respectively, then, calling  the log-probability ``error cost'' $E=d \log \frac{\epsilon}{K}$, we can write the acceptability of $\s'$, $\acc(\s')$, as:
\begin{align}
\acc(\s') \approx W_i \log P(\m^{*}) + W_{ii} \log E.    
\end{align}
As a special case, when the message is grammatically realized, the error distance $E=0$ and $\acc(\s') \approx W_i \log P(\m^{*}) $.

Based on these assumptions, if one compares the acceptabilities of the grammatical string $\s$ and the ungrammatical string $\s'$ in a minimal pair, factor (i) cancels out in the acceptability difference:
\begin{align}
    \acc(\s) &\approx W_i \log P(\m^{*}) \nonumber \\
    \acc(\s') &\approx W_i \log P(\m^{*}) + W_{ii} E \nonumber \\
    \acc(\s) - \acc(\s') &\approx W_{ii} E \label{eq:acc-diff}
\end{align}
and the acceptability difference should correlate with the sentence log-probability difference, $E$, which is Prediction 2a. For pairs that are not meaning-matched, however, we have two different meanings and so factor (i) will not cancel out, weakening the correlation between log-probability difference and acceptability difference. If we further assume that larger differences in message tends to imply larger differences in message log-probability, then correlations between acceptability difference and log-probability difference will drop with larger differences in message, which is Prediction 2b.

\subsection{Prediction 3} \label{sec:derivation-pred3}

\begin{pred}
    Potentially poor separation based on probability between grammatical/ungrammatical strings.
\end{pred}
We consider a pooled set of grammatical and ungrammatical strings. We can measure the ``separation'' based on $P(P(g) > P(u))$, where $g$ is a grammatical string drawn from the pool, $u$ is an ungrammatical string drawn from the pool, and the two draws are independent. Note that this is equivalent to the area under the receiver operating characteristic (ROC) curve for the probabilities assigned to each string in the set. If this quantity is 1, this means that $P(g)$ is always greater than $P(u)$. 

We first analyze the case for minimal pairs, where there is the greatest reason to expect good separation \emph{a priori}. Using similar logic as above, we have:
\begin{align}
    P(g) 
    &\approx (1-\epsilon)P(\m_g)\\
    P(u) 
    &\approx \frac{\epsilon}{K} P(\m_u)
\end{align}
where $\m_g$ is the unique message associated with $g$, and $\m_u$ is the highest probability message given $u$, i.e., $\m_u = \argmax_{\m} P(u | \m) $.

Therefore,
\begin{align}
    \log \frac{P(g)}{P(u)} &\approx \log P(\m_g) - \log P(\m_u) - \log \epsilon + \log K
\end{align}
and so, approximately, $P(g) > P(u)$ iff:
\begin{align}
    \log P(\m_g) - \log P(\m_u) > \log \epsilon - \log K
    \label{eq:final-prediction-3-equation}
\end{align}

When the population of strings consists of minimal pairs, $\m_g$ and $\m_u$ are drawn from the same distribution $P(\m)$, so $\log P(\m_g) - \log P(\m_u)$ is symmetric around 0. Furthermore, since $0 < \epsilon < 1$, and $K \geq 1$, we know that $\log \epsilon - \log K$ must be negative. 
For any symmetric distribution, with $\mu = 0$, and PDF $f$, $\int_{-\infty}^{0}f(x)dx < 0.5$.
Therefore, it is guaranteed that the AUC $P(P(g) > P(u)) > 0.5$. However, its actual value depends greatly on the variance of $\log P(m)$, and might well not be much above 0.5.

This analysis also offers a way of understanding why and under what conditions minimal pairs offer good testing grounds for LMs' grammatical capabilities. ``Better'' LMs can be expected to place higher probability on strings that (i) are grammatical, and (ii) convey plausible meanings. Factor (i) means that ``better'' LMs will have lower values of $\epsilon$; factor (ii) will, if anything, tend to narrow the range of log-probabilities of plausible meanings (pushing them to higher values), and hence reduce the variance of $\log P(\m)$. Among LMs clearing a reasonable minimum bar of quality, $K$ is in contrast best thought of as a property of the grammatical strings in the set of minimal pairs, not as a property differing among LMs, so it is a constant for purposes of this analysis. Therefore, both factors (i) and (ii) will increase AUC.

This analysis can be broadened to populations of strings beyond just sets of minimal pairs: in those cases, $\m_g$ and $\m_u$ are not drawn from the same distribution, so there is even less reason to expect high AUC. Indeed, if $P(\m_u)$ tends to be higher than $P(\m_g)$, we might even see AUC < 0.5.

\paragraph{Generalizing to transformations of string probability.} \label{app:tranformations}
As shown above, for any LM that achieves $1-\epsilon > \frac{\epsilon}{K}$, the limiting factor on AUC is the distributions of $\log P(\m_g)$ and $\log P(\m_u)$. For the case of minimal pairs, 
replacing the two string log-probability terms on the left-hand side of \Cref{eq:final-prediction-3-equation} with random variables that are lower-variance will generally increase AUC: the two terms are drawn from the same distribution, thus their expectation is zero, and thus the lower their variance the greater the probability will tend to be that the inequality is satisfied. 

Two transformations of interest described in \Cref{sec:sentence-scoring-metrics} can be analyzed fairly straightforwardly: dividing string probability by  probability under a uniform per-token distribution, and dividing string probability by its probability under a unigram per-token distribution (our Metrics 2 and 3). These transformations effectively replace LM log-probability with the difference of uniform or unigram log-probability from LM log-probability. In general, for two random variables $X$ and $Y$ with standard deviations $\sigma_X$ and $\sigma_Y$ and correlation $\rho$, we have $\sigma^2_{X-Y}=\sigma^2_{X} + \sigma^2_{Y} - 2\rho \sigma_X\sigma_Y$. Therefore, $\sigma^2_{X-Y} <\sigma^2_{X}$ if $\sigma_Y < 2 \rho \sigma_X$. That is, the random variables must be positively correlated, and $Y$'s variance must not be too large relative to that of $X$. These criteria are good candidates for being met by both transformations. Among grammatical strings, those with lower LM probabilities tend to be longer and haver rarer words, so the distribution of grammatical string probabilities under an LM is likely to be positively correlated with the distributions of grammatical string probabilities under both uniform and unigram distributions. And, since LMs are better models than uniform and unigram models, LM log-probabilities among a grammatical set of strings will tend to have lower variance---clustered more tightly toward 0---than the log-probabilities of uniform or unigram distributions over the same set of grammatical strings.

We have found Metrics 4 and 5 less straightforward to analyze, but speculate that a generally similar approach may be feasible.

No corresponding general results can be offered for analysis of AUC among pools of strings not consisting of sets of minimal pairs, because of the differences in distributions of $P(\m_g)$ and $P(\m_u)$. Indeed, sets of strings adversarial to a metric could be constructed; for example, selectively including only grammatical strings involving frequent words and only ungrammatical strings involving rare words would be adversarial to Metric 3. However, we expect that these approaches will generally tend to improve AUC over that obtained by using raw string probabilities, when sentence sets are not adversarially constructed.

\section{Additional figures for \Cref{pred:no-separation}} \label{sec:appendix-no-separation}

\begin{figure*}[ht]
    \centering
    \subfloat[\gpt]{
        \includegraphics[width=0.7\linewidth]{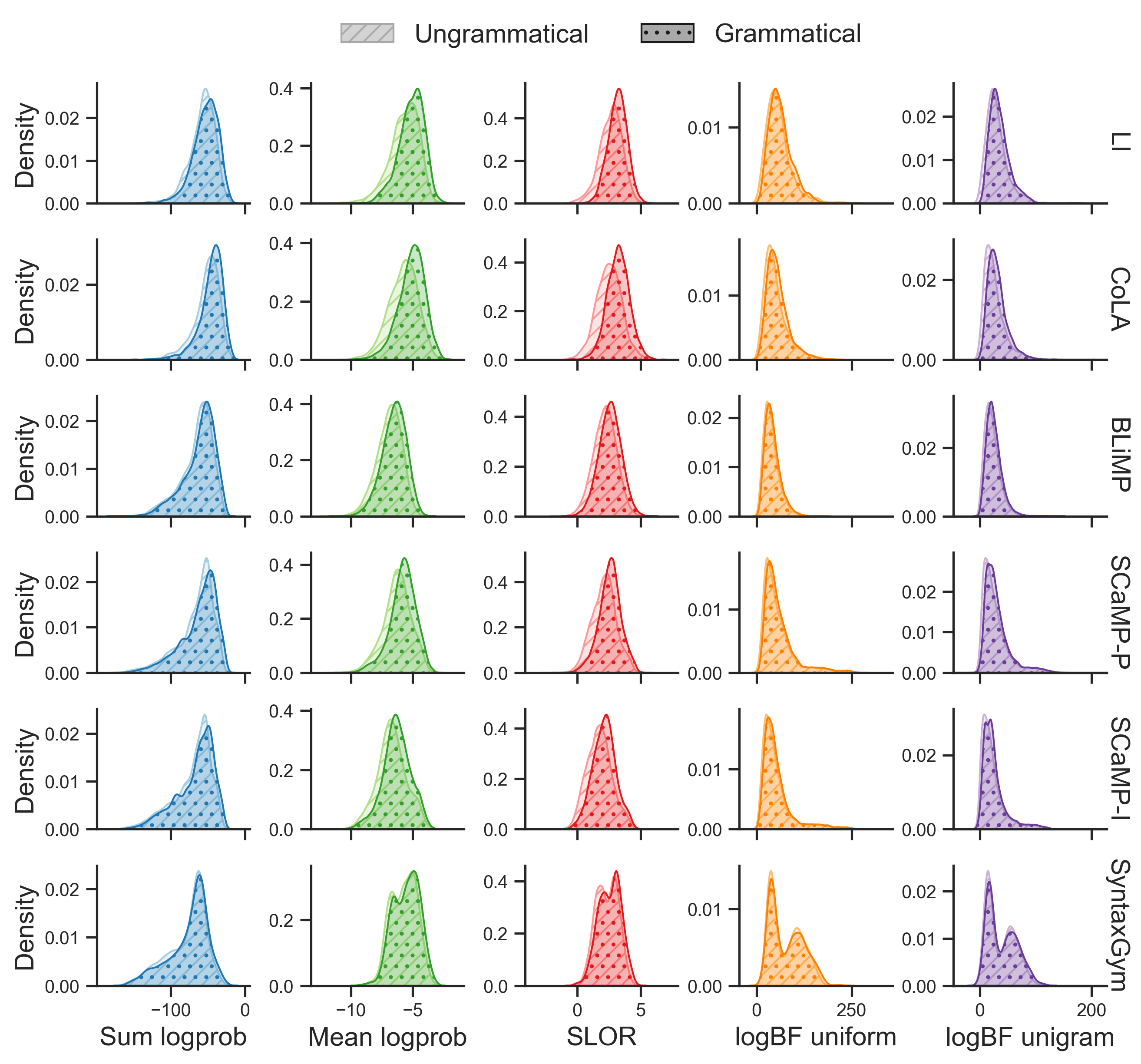}
    }\\
    \subfloat[\llamathree]{
        \includegraphics[width=0.7\linewidth]{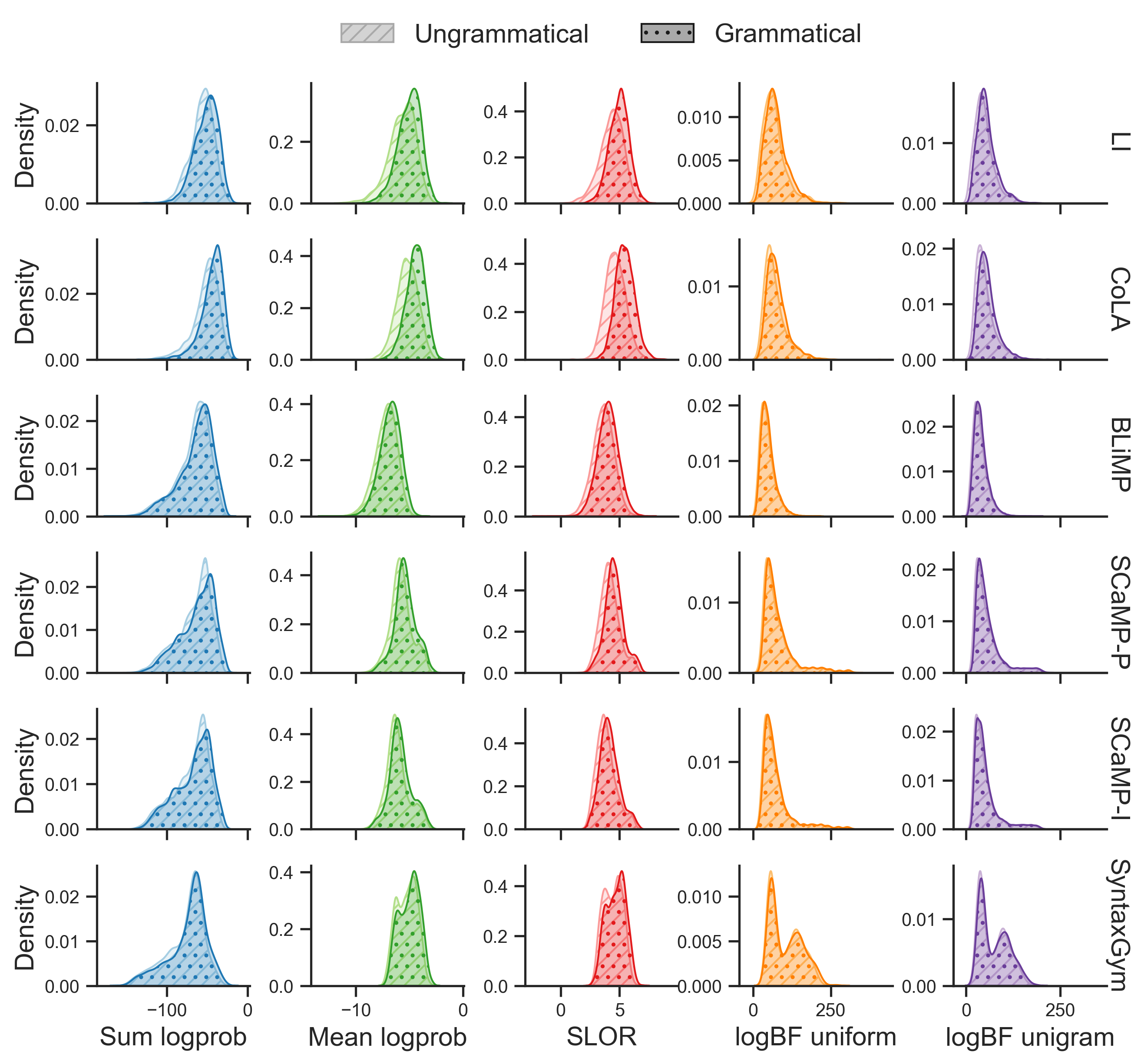}
    }
    \caption{Score distributions for grammatical and ungrammatical sentences from each English dataset.}
    \label{fig:pooled-scores-all-datasets}
\end{figure*}

\begin{figure*}[ht]
    \centering
    \subfloat[\gpt]{
        \includegraphics[width=\linewidth]{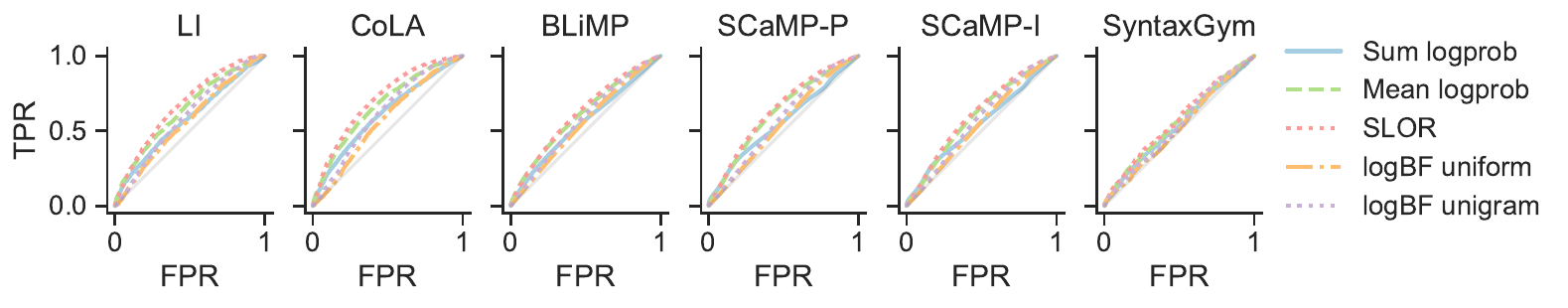}
    }\\
    \subfloat[\llamathree]{
        \includegraphics[width=\linewidth]{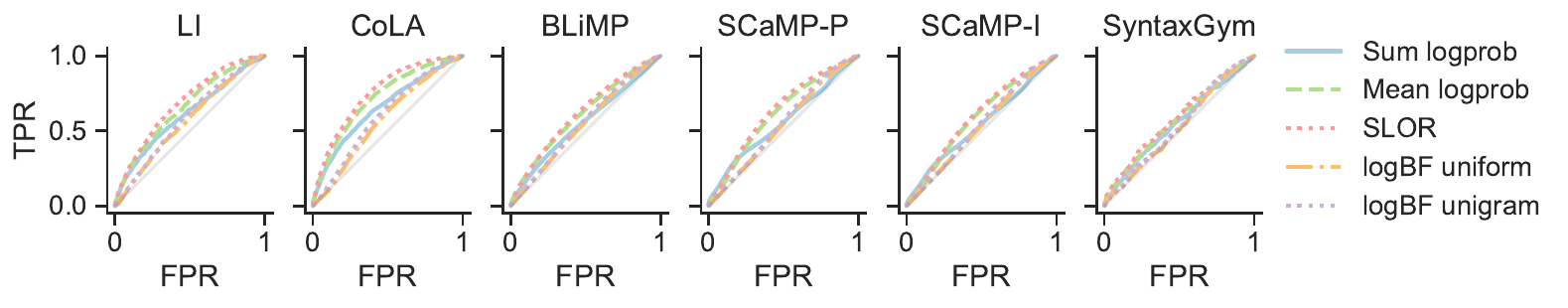}
    }
    \caption{ROC curves achieved by models on each English dataset.}
    \label{fig:roc}
\end{figure*}

\begin{figure*}[ht]
    \centering
    \includegraphics[width=0.7\linewidth]{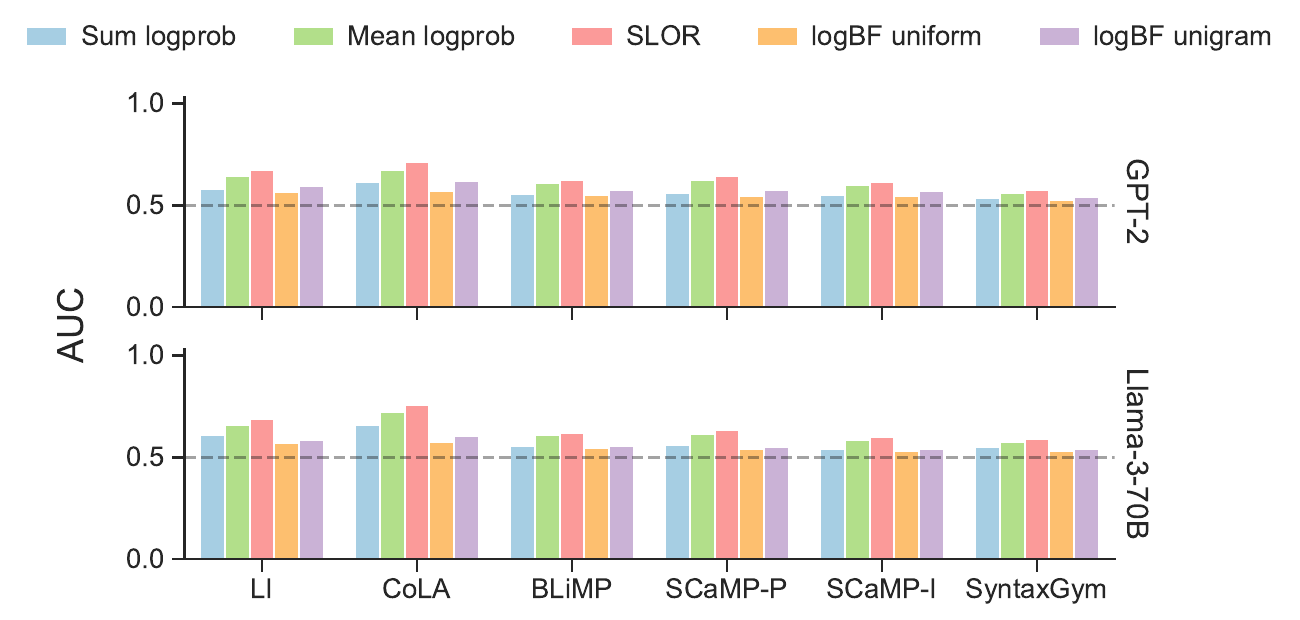}
    \caption{AUC scores for each model (rows), metric (hue), and dataset ($x$-axis), corresponding to ROC curves in \Cref{fig:roc}.}
    \label{fig:auc-all-datasets}
\end{figure*}

\begin{figure*}[ht]
    \centering
    \includegraphics[width=\linewidth]{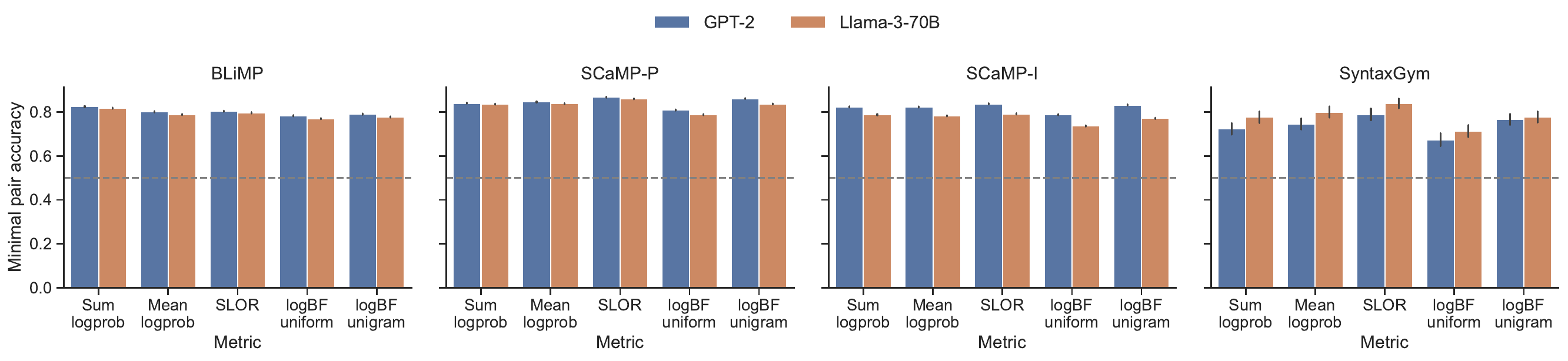}
    \caption{Accuracy (i.e., proportion of pairs where the grammatical sentence receives a higher score) achieved on English minimal pair datasets, using the five scoring functions discussed in \Cref{sec:sentence-scoring-metrics}.}
    \label{fig:minpair-acc}
\end{figure*}